\documentclass{article}

\usepackage[final, nonatbib]{neurips_2021}

\usepackage[utf8]{inputenc} %
\usepackage[T1]{fontenc}    %
\usepackage{hyperref}       %
\usepackage{url}            %
\usepackage{booktabs}       %
\usepackage{amsfonts}       %
\usepackage{nicefrac}       %
\usepackage{microtype}      %
\usepackage{xcolor}         %

\usepackage{graphicx}
\usepackage{multirow}
\usepackage{booktabs}
\usepackage{floatrow}%
\usepackage{blindtext}
\usepackage[export]{adjustbox}
\usepackage[ruled,vlined]{algorithm2e}
\usepackage{amsmath}
\newcommand{\bigCI}{\mathrel{\text{\scalebox{1.07}{$\perp\mkern-10mu\perp$}}}}

\usepackage{caption}
\usepackage{subcaption}
\newfloatcommand{capttabbox}{table}[\captop][\FBwidth ]

\title{Spatial-Temporal Super-Resolution of Satellite Imagery via Conditional Pixel Synthesis}

\author{%
  Yutong He
  \qquad Dingjie Wang \qquad Nicholas Lai \qquad William Zhang \qquad Chenlin Meng \\ \textbf{Marshall Burke \qquad David B. Lobell \qquad Stefano Ermon}\\
    Stanford University \\
  \{\tt kellyyhe, daviddw, nicklai, wxyz, chenlin, ermon\}@cs.stanford.edu \\
  \{\tt mburke, dlobell\}@stanford.edu \\
}

\begin{document}
\maketitle

\begin{abstract}
\label{abstract}

High-resolution satellite imagery has proven useful for a broad range of tasks, including measurement of global human population, local economic livelihoods, and biodiversity, among many others. Unfortunately, high-resolution imagery is both infrequently collected and expensive to purchase, making it hard to efficiently and effectively scale these downstream tasks over both time and space. We propose a new conditional pixel synthesis model 
that uses abundant, low-cost, low-resolution imagery 
to generate accurate high-resolution imagery at locations and times in which it is unavailable.  
We show that our model attains photo-realistic sample quality and outperforms competing baselines on a key downstream task -- object counting -- particularly in geographic locations where conditions on the ground are changing rapidly.

\end{abstract}

\section{Introduction}
\label{intro}

Recent advancements in satellite technology have enabled granular insight into the evolution of human activity on the planet's surface. Multiple satellite sensors now collect imagery with spatial resolution 
less than 1m, and this high-resolution (HR) imagery can provide sufficient information for various fine-grained tasks such as 
post-disaster building damage estimation, poverty prediction, and crop phenotyping \cite{gupta2019xbd, ayush2020generating, ZHANG2020105584}. Unfortunately, HR imagery is captured infrequently over much of the planet's surface (once a year or less), especially in developing countries where it is arguably most needed, and was historically captured even more rarely (once or twice a decade) \cite{burke2021using}. Even when available, HR imagery is prohibitively expensive to purchase in large quantities. These limitations often result in an inability to scale promising HR algorithms and apply them to questions of broad social importance. 
Meanwhile, multiple sources of publicly-available satellite imagery now provide sub-weekly coverage at global scale, albeit at lower spatial resolution (e.g. 10m resolution for Sentinel-2). Unfortunately, such coarse spatial resolution renders small objects like residential buildings, swimming pools, and cars unrecognizable.

\begin{figure}[h!]
    \centering
    \includegraphics[width = 0.85\textwidth]{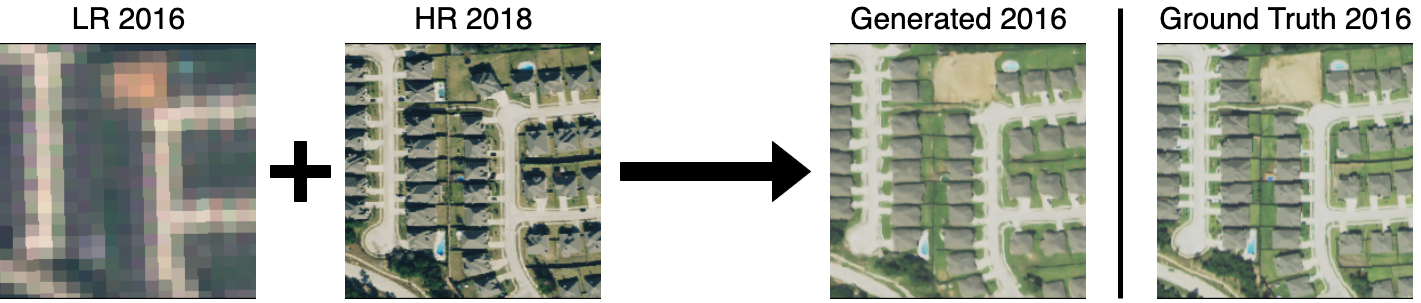}
    \caption{Given a 10m low resolution (LR) image from 2016 and a 1m high resolution (HR) image from 2018, we generate a photo-realistic and accurate HR image for 2016.}
    \label{fig:intro}
\end{figure}

In the last few years, thanks to advances in deep learning and generative models, we have seen great progress in image processing tasks such as image colorization \cite{colorization}, denoising \cite{denoising1, denoising2}, inpainting \cite{denoising2, inpainting}, and super-resolution \cite{srcnn, srgan, dbpn}.
Furthermore, pixel synthesis models such as neural radiance field (NeRF) \cite{nerf} have demonstrated great potential for generating realistic and accurate scenes from different viewpoints. 
Motivated by these successes and the need for high-resolution images, we ask whether it is possible to synthesize high-resolution satellite images using deep generative models. For a given time and location, can we  generate a  high-resolution image  by interpolating the available low-resolution and high-resolution images collected over time?

To address this question, we propose a conditional pixel synthesis model that leverages the fine-grained spatial information in HR images and the abundant temporal availability of LR images to create the desired synthetic HR images of the target location and time. Inspired by the recent development of pixel synthesis models pioneered by the NeRF model~\cite{nerf, pixelnerf, cips}, each pixel in the output images is generated conditionally independently
by a perceptron-based generator given the encoded input image features associated with the pixel, the positional embedding of its spatial-temporal coordinates, and a random vector. Instead of learning to adapt to different viewing directions in a single 3D scene~\cite{nerf}, our model learns to interpolate across the time dimension for different geo-locations with the two multi-resolution satellite image time series.

To demonstrate the effectiveness of our model, we collect a large-scale paired satellite image dataset of residential neighborhoods in Texas using high-resolution NAIP (National Agriculture Imagery Program, 1m GSD) and low-resolution Sentinel-2 (10m GSD) imagery. This dataset consists of scenes in which housing construction occurred between 2014 and 2017 in  major metropolitan areas of Texas, with construction verified using CoreLogic tax and deed data. These scenes thus provide a rapidly changing environment on which to assess model performance. As a separate test, we also pair HR images (0.3m to 1m GSD) from the Functional Map of the World (fMoW) dataset \cite{christie2018functional} crop field category with images from Sentinel-2. 

To evaluate our model's performance, we 
compare 
to 
state-of-the-art methods, including 
super-resolution models. Our model outperforms all competing models in sample quality on both datasets measured by both standard image quality assessment metrics and human perception (see example in Figure ~\ref{fig:intro}). 
Our model also achieves $0.92$ and $0.62$ Pearson's $r^2$ in reconstructing the correct numbers of buildings and swimming pools respectively in the images, outperforming other models in these tasks. Results suggest our model's potential to scale to downstream tasks that use these object counts as input, including societally-important tasks such as population measurement, poverty prediction, and humanitarian assessment \cite{burke2021using,ayush2020generating}.

\section{Related Work}

\paragraph{Image Super-resolution}
SRCNN \cite{srcnn} is the first paper to introduce convolutional layers into a SR context and demonstrate significant improvement over traditional SR models.
SRGAN \cite{srgan} improves on SRCNN with adversarial loss and is widely compared among many GAN-based SR models for remote sensing imagery \cite{udgan, drgan, re-esrgan}. DBPN \cite{dbpn} is a state-of-the-art SR solution that uses an iterative algorithm to provide an error feedback system, and it is one of the most effective SR models for satellite imagery \cite{hybrid}. However, \cite{effectofsronsatellite} shows that SR is less beneficial at coarser resolution, especially when applied to downstream object detection on satellite imagery. In addition, most SR models test on benchmarks where LR images are artificially created, instead of collected from actual LR devices \cite{div2k, benchmark, urban100}. 
SR models also generally perform worse at larger scale factors, which is closer to settings for satellite imagery SR in real life.%

SRNTT \cite{zhang2019image} applies reference-based super-resolution through neural texture transfer to mitigate information loss in LR images by leveraging texture details from HR reference images. While SRNTT also uses a HR reference image, it does not learn the additional time dimension to leverage the HR image of the same object at a different time. In addition, our model uses a perceptron based generator while SRNTT uses a CNN based generator.  

\paragraph{Fusion Models for Satellite Imagery}
\cite{starfm} first proposes STARFM to blend data from two remote sensing devices, MODIS \cite{modis} and Landsat \cite{landsat}, for spatial-temporal super resolution to predict land reflectance. \cite{estarfm} introduces an enhanced algorithm for the same task and \cite{unmixing} combines linear pixel unmixing and STARFM to improve spatial details in the generated images. cGAN Fusion \cite{cganfusion} incorporates GAN-based models in the solution, using an architecture similar to Pix2Pix \cite{pix2pix}. In contrast to previous work, we are particularly interested in synthesizing images with very high resolution ($\leq$ 1m GSD), enabling downstream applications such as poverty level estimation.%

\paragraph{NeRF and Pixel Synthesis Models}
Recent developments in deep generative models, especially advances in perceptron-based generators, have yet to be explored in remote sensing applications. Introduced by \cite{nerf}, neural radiance fields (NeRF) demonstrates great success in constructing 3D static scenes. \cite{nsff, spacetimenerf} extends the notion of NeRF and incorporates time-variant representations of the 3D scenes. \cite{graf} embeds NeRF generation into a 3D aware image generator. These works, however, are limited to generating individual scenes, in contrast with our model which can generalize to different locations in the dataset. \cite{pixelnerf} proposes a framework that predicts NeRF conditioning on spatial features from input images; however, it requires constructing the 3D scenes, which is less applicable to satellite imagery. \cite{cips} proposes a style-based 2D image generative model using an only perceptron-based architecture; however, unlike our method, it doesn't consider the task of conditional 2D image generation nor does it incorporate other variables such as time. In contrast, we propose a pixel synthesis model that learns a conditional 2D spatial coordinate grid along with a continuous time dimension, which is tailored for remote sensing, where the same location can be captured by different devices (e.g. NAIP or Sentinel-2) at different times (e.g. year 2016 or year 2018).

\section{Problem Setup}
The goal of this work is to develop a method to synthesize high-resolution satellite images for locations and times for which these images are not available. As input we are given two time-series of high-resolution (HR) and low-resolution (LR) images for the same location. 
Intuitively, we wish to leverage the rich information in HR images and the high temporal frequency of LR images to achieve the best of both worlds.

Formally, let $I_{hr}^{(t)} \in \mathbf{R}^{C \times H \times W}$ be a sequence of random variables representing high-resolution views of a location at various time steps $t \in T$. Similarly, let $I_{lr}^{(t)}\in \mathbf{R}^{C \times H_{lr} \times W_{lr}}$ denote low-resolution views of the same location over time. Our goal is to develop a method to estimate $I_{hr}^{(t)}$, given $K$ high resolution observations $\{I_{hr}^{(t'_1)}, \cdots, I_{hr}^{(t'_K)}\}$ and $L$ low-resolution ones  $\{I_{lr}^{(t''_1)}, \cdots, I_{lr}^{(t''_L)}\}$ for the same location.
Note the available observations could be taken either before or after the target time $t$. Our task can be viewed as a special case of multivariate time-series imputation, where two concurrent but incomplete series of satellite images of the same location in different resolutions are given, and the model should predict the most likely pixel values at an unseen time step of one image series.

In this paper, we consider a special case where the goal is to estimate $I_{hr}^{(t)}$ given a single high-resolution image $I_{hr}^{(t')}$ and a single low-resolution image $I_{lr}^{(t)}$ also from time $t$.
We focus on this special case because while typically $L \gg K$, it is reasonable to assume $I_{hr}^{(t)} \bigCI I_{lr}^{(t')} \mid I_{lr}^{(t)}$ for $t' \neq t$, i.e., given a LR image at the target time $t$, other LR views from different time steps provide little or no additional information. 
Given the abundant availability of LR imagery, it is often safe to assume access to $I_{lr}^{(t)}$ at target time $t$. Figure \ref{fig:intro} provides a visualization of this task. 

For training, we assume access to paired triplets $\{I_{hr}^{(t)}, I_{lr}^{(t)}, I_{hr}^{(t')}\}$ collected across a geographic region of interest where $t' \neq t$. At inference time, we assume availability for $I_{lr}^{(t)}$ and $I_{hr}^{(t')}$ and the model needs to generalize to previously unseen locations. Note that at inference time, the target time $t$ and reference time $t'$ may not have been seen in the training set either.

\section{Method}
\label{method}

\begin{figure}[t]
    \centering
    \includegraphics[width = \textwidth]{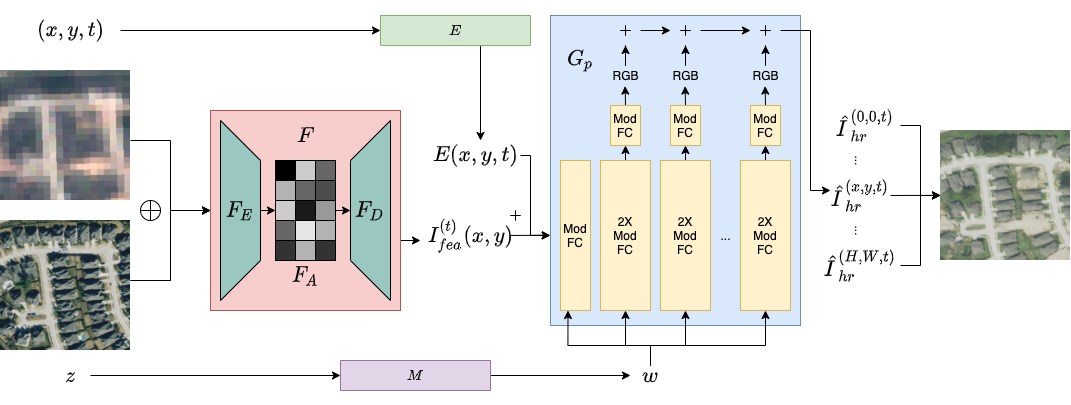}
    \caption{An illustration of our proposed framework (discriminator omitted). The input images are processed by the image feature mapper $F$ to obtain $I_{fea}^{(t)}$. Then with its spatial-temporal coordinate $(x,y,t)$ encoded by $E$, each pixel is synthesized conditionally independently given the image feature associated with its spatial coordinate $I_{fea}^{(t)}(x,y)$ and a random vector $z$.}
    \label{fig:model}
\end{figure}

Given $I_{lr}^{(t)}$ and $I_{hr}^{(t')}$ of the target location and target time $t$, our method generates $\hat{I}_{hr}^{(t)} \in \mathbf{R}^{C \times H \times W}$ with a four-module conditional pixel synthesis model. Figure \ref{fig:model} is an illustration of our framework.

The generator $G$ of our model consists of three parts: image feature mapper $F: \mathbf{R}^{C \times H \times W} \to \mathbf{R}^{C_{fea} \times H \times W}$, positional encoder $E$, and the pixel synthesizer $G_{p}$. For each spatial coordinate $(x, y)$ of the target HR image, the image feature mapper extracts the neighborhood information around $(x,y) \in \{0,1,...,H\} \times \{0,1,...,W\}$ from $I_{lr}^{(t)}$ and $I_{hr}^{(t')}$, as well as the global information associated with the coordinate in the two input images. The positional encoder learns a representation of the spatial-temporal coordinate $(x, y, t)$, where $t$ is the temporal coordinate of the target image. The pixel synthesizer then uses the information obtained from the image feature mapper and the positional encoding to predict the pixel value at each coordinate. Finally, we incorporate an adversarial loss in our training, and thus include a discriminator $D$ as the final component of our model.

\paragraph{Image Feature Mapper}
Before extracting features, we first perform nearest neighbor resampling to the LR image $I_{lr}^{(t)}$ to match the dimensionality of the HR image and concatenate $I_{lr}^{(t)}$ and $I_{hr}^{(t')}$ along the spectral bands to form the input $I_{cat}^{(t)} = \textrm{concat}[I_{lr}^{(t)}, I_{hr}^{(t')}] \in \mathbf{R}^{2C \times H \times W}$. Then the mapper processes $I_{cat}^{(t)}$ with a neighborhood encoder $F_E: \mathbf{R}^{2C \times H \times W} \to \mathbf{R}^{C_{fea} \times H' \times W'}$, a global encoder $F_A: \mathbf{R}^{C_{fea} \times H' \times W'} \to \mathbf{R}^{C_{fea} \times H' \times W'}$ and a neighborhood decoder $F_D: \mathbf{R}^{C_{fea} \times H' \times W'} \to \mathbf{R}^{C_{fea} \times H \times W}$. The neighborhood encoder and decoder learn the fine structural features of the images, and the global encoder learns the overall inter-pixel relationships as it observes the entire image.

$F_E$ uses sliding window filters to map a small neighborhood of each coordinate into a value stored in the neighborhood feature map $I_{ne}^{(t)} \in \mathbf{R}^{C_{fea} \times H' \times W'}$ and $F_D$ uses another set of filters to transform the global feature map $I_{gl}^{(t)} \in \mathbf{R}^{C_{fea} \times H' \times W'}$ back to the original coordinate grid. $F_A$ is a self-attention module that takes $I_{ne}^{(t)}$ as the input and learns functions $Q, K: \mathbf{R}^{C_{fea} \times H' \times W'} \to \mathbf{R}^{C_{fea}/8 \times H'W'}, V: \mathbf{R}^{C_{fea} \times H' \times W'} \to \mathbf{R}^{C_{fea} \times H'W'}$ and a scalar parameter $\gamma$ to map $I_{ne}^{(t)}$ to $I_{gl}^{(t)}$.

The image feature mapper $F = F_E \circ F_A \circ F_D$ and we denote $I_{fea}^{(t)} = F(I_{cat}^{(t)})$ and the image feature associated with coordinate $(x,y)$ as $I_{fea}^{(t)}(x,y) \in \mathbf{R}^{C_{fea}}$. Details are available in Appendix A.

\paragraph{Positional Encoder}

Following \cite{cips}, we also include both the Fourier feature and the spatial coordinate embedding in the positional encoder $E$. The Fourier feature is calculated as $e_{f_o}(x,y,t) = \sin(B_{f_o}(\frac{2x}{H-1}-1, \frac{2y}{H-1}-1, \frac{t}{u}))$ where $B_{f_o} \in \mathbf{R}^{3 \times C_{fea}}$ is a learnable matrix and $u$ is the time unit. This encoding of $t$ allows our model to handle time-series with various lengths and to extrapolate to time steps that are not seen at training time. $E$ also learns a $C_{fea}\times H \times W$ matrix $e_{co}$ and the spatial coordinate embedding for $(x,y,t)$ is extracted from the vector at $(x,y)$ in $e_{co}$. The positional encoding of $(x,y,t)$ is the channel concatenation of $e_{f_o}(x,y,t)$ and $e_{co}(x,y,t)$, $E(x,y,t) = \textrm{concat}[e_{f_o}(x,y,t), e_{co}(x,y,t)] \in \mathbf{R}^{2C_{fea}}$.

\paragraph{Pixel Synthesizer}

Pixel Synthesizer $G_p$ can be viewed as an analogy of simulating a conditional $2+1D$ neural radiance field with fixed viewing direction and camera ray using a perceptron based model. Instead of learning the breadth representation of the location, $G_p$ learns to scale in the time dimension in a fixed spatial coordinate grid.
Each pixel is synthesized conditionally independently given $I_{fea}^{(t)}$, $E(x,y,t)$, and a random vector $z \in \mathbf{R}^Z$. $G_p$ first learns a function $g_z$ to map $E(x,y,t)$ to $\mathbf{R}^{C_{fea}}$, then obtains the input to the fully-connected layers $e(x,y,t) = g_z(E(x,y,t)) + I_{fea}^{(t)}(x,y)$. Following \cite{cips, styleganv2}, we use a $m$-layer perceptron based mapping network $M$ to map the noise vector $z$ into a style vector $w$, and use $n$ modulated fully-connected layers (ModFC) to inject the style vector into the generation to maintain style consistency among different pixels of the same image. We map the intermediate features to the output space for every two layers and accumulate the output values as the final pixel output.

With all components combined, the generated pixel value at $(x, y, t)$ can be calculated as \[\hat{I}_{hr}^{(t)}(x,y) = G(x,y,t,z|I_{lr}^{(t)}, I_{hr}^{(t')}) = G_p(E(x,y,t), F(I_{cat}^{(t)}), z)\]

\paragraph{Loss Function}
The generator is trained with the combination of the conditional GAN loss and $L_1$ loss.
The objective function is \[G^* = \arg\min_G\max_D \mathcal{L}_{cGAN}(G,D) + \lambda \mathcal{L}_{L_1}(G)\]
$\mathcal{L}_{cGAN}(G,D) = \mathbb{E}[\log D(I_{hr}^{(t)}, X, I_{lr}^{(t)}, I_{hr}^{(t')})] + \mathbb{E}[1-\log D(G(X, z | I_{lr}^{(t)}, I_{hr}^{(t')}), X, I_{lr}^{(t)}, I_{hr}^{(t')})]$ where $X$ is the temporal-spatial coordinate grid $\{(x,y,t) | 0 \leq x \leq H, 0 \leq y \leq W\}$ for $I_{hr}^{(t)}$. $\mathcal{L}_{L_1}(G) = \mathbb{E}[||I_{hr}^{(t)} - G(X, z | I_{lr}^{(t)}, I_{hr}^{(t')})||_1]$.
\section{Experiments}
\label{experiment}

\subsection{Datasets}
\paragraph{Texas Housing Dataset}

We collect a dataset consisting of 286717 houses and their surrounding neighborhoods from CoreLogic tax and deed database that have an effective year built between 2014 and 2017 in Texas, US. We reserve 14101 houses from 20 randomly selected zip codes as the testing set and use the remaining 272616 houses from the other 759 zip codes as the training set. For each house in the dataset, we obtain two LR-HR image pairs, one from 2016 and another from 2018. In total, there are 1146868 multi-resolution images collected from different sensors for our experiments. We source high resolution images from NAIP (1m GSD) and low resolution images from Sentinel-2 (10m GSD) and only extract RGB bands from Google Earth Engine \cite{gorelick2017google}.
More details can be found in Appendix C.

\paragraph{FMoW-Sentinel2 Crop Field Dataset}
We derive this dataset from the crop field category in Functional Map of the World (fMoW) dataset \cite{christie2018functional} for the task of generating images over a greater number of time steps.
We pair each fMoW image with a lower resolution Sentinel-2 RGB image captured at the same location and a similar time. We prune locations with fewer than 2 timestamps, yielding 1752 locations and a total of 4898 fMoW-Sentinel2 pairs. Each location contains between 2-15 timestamps spanning from 2015 to 2017. We reserve 237 locations as the testing set and the remaining 1515 locations as the training set. More details can be found in Appendix C.

\subsection{Implementation Details}
\paragraph{Model Details}
We choose $H = W = 256$, $C = 3$ (the concatenated RGB bands of the input images), $C_{fea} = 256$, $m = 3$, $n = 14$ and $\lambda = 100$ for all of our experiments. We use non-saturating conditional GAN loss for $G$ and $R_1$ penalty for $D$, which has the same network structure as the discriminator in \cite{styleganv2, cips}. We train all models using Adam optimizer with learning rate $2\times 10^{-3}, \beta_0 = 0, \beta_1 = 0.99, \epsilon = 10^{-8}$.
We train each model to convergence, which takes around 4-5 days on 1 NVIDIA Titan XP GPU.
Further details can be found in Appendix A and B.

We provide two versions of the image feature mapper. In version "EAD", we use convolutional layers and transpose convolutional layers with stride $> 1$ in $F_E$ and $F_D$. In version "EA", we use convolutional layers with stride $=1$ in $F_E$ and an identity function in $F_D$. For version "EA", we use a patch-based training and inference method with a patch size of $64$ because of memory constraints, and denote it as "EA64". The motivation for including both "EAD" and "EA" is to examine the capabilities of $F$ with and without spatial downsampling or upsampling. "EAD" can sample 1500 images in around 2.5 minutes (10 images/s) and "EA64" can sample 1500 images in around 19 minutes (1.3 image/s). More details can be found in Appendix A and B.

\paragraph{Baselines} We compare our method with two groups of baseline methods: image fusion models and super-resolution (SR) models. 
We use cGAN Fusion \cite{cganfusion}, which leverages the network structure of the leading image-to-image translation model Pix2Pix \cite{pix2pix} to combine different imagery products for surface reflectance prediction. We also compare our model with the original Pix2Pix framework.
For SR baselines, we choose SRGAN \cite{srgan}, which is widely compared among other GAN based SR models for satellite imagery \cite{udgan, drgan, re-esrgan}. We also compare our method with DBPN \cite{dbpn}, which is a state-of-the-art SR model for satellite imagery \cite{hybrid}.

\subsection{Image Generation Quality}

\begin{figure}[h!]
    \centering
    \includegraphics[width = \textwidth]{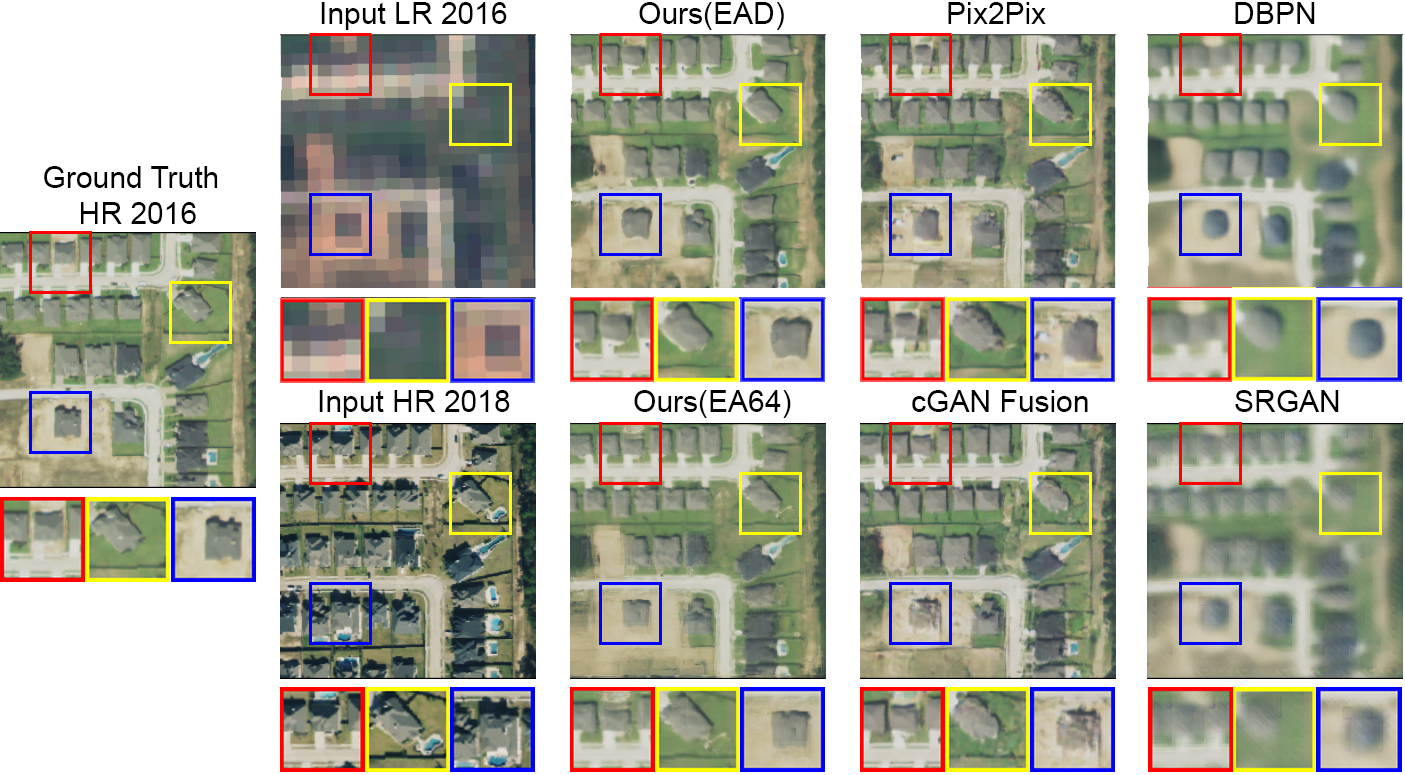}
    \caption{Samples from all models on the Texas housing dataset with setting $t' > t$.
    Our models show advantages in both sample quality and structural detail consistency with the ground truth, especially in areas with house or pool construction (zoomed in with colored boxes).
    }
    \label{fig:housing_sample}
\end{figure}

\begin{figure}[h!]
    \centering
    \includegraphics[width = \textwidth]{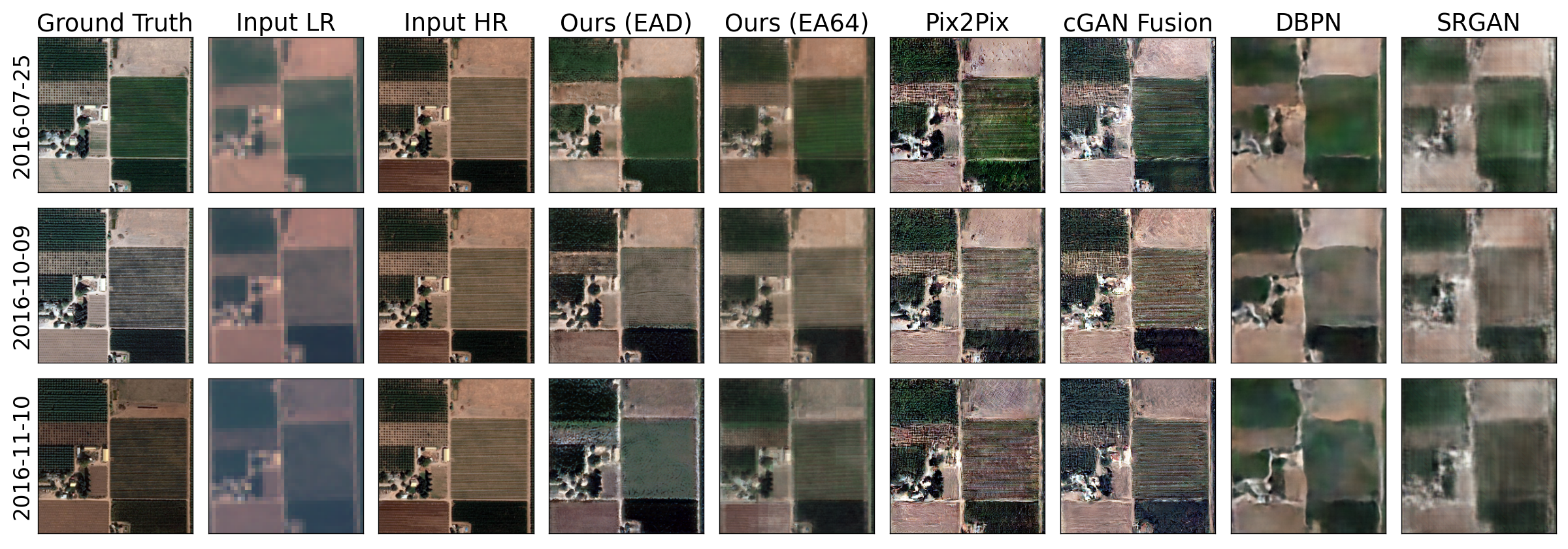}
    \caption{Samples from all models on the fMoW-Sentinel2 crop field dataset. Each row represents the results on the same location at a different timestamp given the same HR input from 2016-09-06.}
    \label{fig:fmow_sample}
\end{figure}

\begin{figure}[h!]
\CenterFloatBoxes
\begin{floatrow}
\ffigbox[\FBwidth]{%
  \includegraphics[width = 0.39\textwidth]{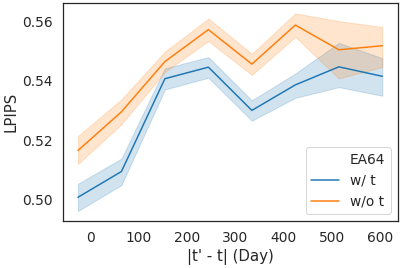}
}{%
  \caption{Ablation study on learning the temporal embeddings in our model using fMoW-Sentinel2 crop field dataset.}%
  \label{fig:t_ablation}
}
\capttabbox{%
    \begin{tabular}{c|cccc} 
    \toprule
    Model       & SSIM$\uparrow$            & PSNR$\uparrow$             & FSIM$\uparrow$            & LPIPS$\downarrow$            \\ 
    \hline
    Pix2Pix     & 0.2144          & 14.0276          & 0.6418          & 0.5847           \\
    cGAN Fusion & 0.2057          & 14.1353          & 0.6409          & 0.5912           \\
    DBPN        & 0.3621          & 15.7878          & 0.6323          & 0.6428           \\
    SRGAN       & 0.3479          & 15.3502          & 0.6323          & 0.6301           \\ 
    \hline
    Ours (EAD)  & 0.3526          & 16.5769          & \textbf{0.6887} & 0.5629           \\
    Ours (EA64) & \textbf{0.3905} & \textbf{16.8879} & 0.6827          & \textbf{0.5197}  \\
    \bottomrule
    \end{tabular}
}{%
  \caption{Image sample quality quantitative results on fMoW-Sentinel2 crop field dataset.}%
  \label{tab:fmow}
}
\end{floatrow}
\end{figure}

We examine generated image quality using both our Texas housing dataset and our fMoW-Sentinel2 crop field dataset. Figures \ref{fig:housing_sample} and \ref{fig:fmow_sample} present qualitative results from our approach and from baselines. Table \ref{tab:housing} shows quantitative results on the Texas housing dataset and Table \ref{tab:fmow} shows quantitative results on the fMoW-Sentinel2 crop field dataset. Overall, our models outperform all baseline approaches in all evaluation metrics.

\paragraph{Evaluation Metrics} To assess image generation quality, we report standard sample quality metrics SSIM \cite{1284395}, FSIM \cite{5705575}, and PSNR to quantify the visual similarity and pixel value accuracy of the generated images. We also include LPIPS \cite{zhang2018perceptual} using VGG \cite{simonyan2015deep} features, which is a deep perceptual metric used in previous works to evaluate generation quality in satellite imagery \cite{use_lpips}. LPIPS leverages visual features learned in deep convolutional neural networks that better reflect human perception. We report the average score of each metric given every possible pair of $t, t'$ where $t \neq t'$.

\begin{table}[h!]
\centering
\begin{tabular}{c|cccc|cccc} 
\toprule
\multirow{2}{*}{Model} & \multicolumn{4}{c|}{$t' > t$}                                  & \multicolumn{4}{c}{$t' < t$}                                  \\ 
\cline{2-9}
                       & SSIM$\uparrow$           & PSNR$\uparrow$             & FSIM$\uparrow$            & LPIPS$\downarrow$       & SSIM$\uparrow$            & PSNR$\uparrow$             & FSIM$\uparrow$           & LPIPS$\downarrow$        \\ 
\hline
Pix2Pix                & 0.5432         & 20.8420          & 0.7522          & 0.4243          & 0.3909          & 17.9528          & 0.6802         & 0.4909           \\
cGAN Fusion            & 0.5976         & 21.5226          & 0.7713          & 0.3936          & 0.4220          & 17.8763          & 0.6897         & 0.4726           \\
DBPN                   & 0.5781         & 21.4716          & 0.7102          & 0.5101          & 0.4572          & 18.9330          & 0.6384         & 0.5910           \\
SRGAN                  & 0.5361         & 21.1968          & 0.6999          & 0.5261          & 0.4221          & 18.9772          & 0.6387         & 0.5694           \\ 
\hline
Ours (EAD)           & 0.6470         & 22.4906          & \textbf{0.7904} & \textbf{0.3695} & 0.5225          & 19.7675          & \textbf{0.7280} & \textbf{0.4275}  \\
Ours (EA64)           & \textbf{0.6570} & \textbf{22.5552} & 0.7902          & 0.3764          & \textbf{0.5338} & \textbf{19.8547} & 0.7269         & 0.4342           \\
\bottomrule
\end{tabular}
\caption{Image sample quality quantitative results on Texas housing data. $t' > t$ denotes the task for generating an image in the past given a future HR image, and $t' < t$ denotes the task for generating an image in the future given a past HR image.}
\label{tab:housing}
\end{table}

\paragraph{Texas Housing Dataset} The dataset focuses on regions with residential building construction between 2014 and 2017, so we separately analyze the task to predict $\hat{I}_{hr}^{(t)}$ when $t' > t$ and when $t' < t$ on the Texas housing dataset. $t' > t$ represents the task to "rewind" the construction of the neighborhood and $t' < t$ represents the task to "predict" construction. Our models achieve more photo-realistic sample quality (measured by LPIPS), maintain better structural similarity (measured by SSIM and FSIM), and obtain higher pixel accuracy (measured by PSNR) in both tasks compared to other approaches. With the more challenging task $t' < t$, where input HR images contain fewer constructed buildings than the ground truth HR images, our method exceeds the performance of other models by a greater margin.

Our results in Figure \ref{fig:housing_sample} confirm these findings with qualitative examples. Patches selected and zoomed in with colored bounding boxes show regions with significant change between $t$ and $t'$. Our model generates images with greater realism and higher structural information accuracy compared to baselines. While Pix2Pix and cGAN Fusion are also capable of synthesizing convincing images, they generate inconsistent building shapes, visual artifacts, and imaginary details like the swimming pool in the red bounding boxes. DBPN and SRGAN are faithful to information provided by the LR input but produce blurry images that are far from the ground truth.

\paragraph{fMoW-Sentinel2 Crop Field Dataset} We conduct experiments on the fMoW-Sentinel2 crop field dataset to compare model performance in settings with less data, fewer structural changes, and longer time series with unseen timestamps at test time.
Our model outperforms baselines in all metrics, see Table \ref{tab:fmow}. Figure \ref{fig:fmow_sample} shows the image samples from different models on the fMoW-Sentinel2 crop field dataset. While image-to-image translation models fail to maintain structural similarity and SR models fail to attain realistic details, our model generates precise and realistic images.

\paragraph{Discussion}
It is not surprising that our models generate high resolution details because they leverage a rich prior of what HR images look like, acquired via the cGAN loss, and GANs are capable of learning to generate high frequency details \cite{styleganv2, cips}. Despite considerable information loss, inputs from LR devices still provide sufficient signal for HR image generation (e.g. swimming pools may change LR pixel values in a way that is detectable by our models but not by human perception). Experiments in Figure \ref{fig:housing_sample} and Section \ref{human_eval} show that these signals are enough for our model to reliably generate HR images that are high quality and applicable to downstream tasks.

In more extreme scenarios (e.g. LR captured by MODIS with 250m GSD v.s. HR captured by NAIP with 1m GSD), LR provides very limited information and therefore yields excessive uncertainty in generation. In this case, the high resolution details generated by our model are more likely to deviate from the ground truth.

It is worth noting that our LR images are captured by remote sensing devices (e.g. Sentinel-2), as opposed to synthetic LR images created by downsampling used in many SR benchmarks. As shown in our experiments, leading SR models such as DBPN and SRGAN do not perform well in this setting.

\subsection{Ablation Study}
\label{ablation_study}
We perform an ablation study on different components of our framework.
We consider the following configurations for comparison: "No $G_P$" setting removes the pixel synthesizer to examine the effects of $G_P$; "Linear $F$" and "E only" use a single fully-connected layer and a single $3 \times 3$ convolutional layer with stride $=1$ respectively to verify the influence of a deep multi-layer image feature mapper $F$. "ED Only" removes the global encoder $F_A$ and "A Only" removes the neighborhood encoder $F_E$ and decoder $F_D$. Note that because \cite{cips} has conducted thorough analysis on various settings of the positional encoder, we omit the configurations to assess the effects of spatial encoding in $E$.

As shown in Table \ref{tab:ablation}, each component contributes significantly to performance in all evaluation aspects. While "EA64" outperforms "EAD" in SSIM and PSNR with a small margin, we observe slight checkerboard artifacts in the images generated by "EA64"
(details in Appendix G). Overall, "EAD" is the most realistic to the human eye, which is consistent with the LPIPS results. However, "EA64" has stronger performance in a more data-constrained setting as shown in the fMoW-Sentinel2 crop field experiment. Samples generated by different configurations can be found in Appendix F.

We also demonstrate the effects of learning the time dimension in our model. Parameterizing our model with a continuous time dimension enables it to be applicable to time series of varying lengths with non-uniform time intervals (e.g. fMoW-Sentinel2 Crop Field dataset). Moreover, this parameterization also improves model performance. We train a modified version of "EA64" to exclude the time component in $E$ and compare the LPIPS values of the generated images to ones from the original "EA64" in Figure \ref{fig:t_ablation}. In conjunction with the additional analysis in Appendix F, we show that the time-embedded model outperforms the same model without temporal encoding.
Therefore, the time dimension is crucial to our model's performance, especially in areas with sparse HR satellite images over long periods of time.

We also include further details and analysis of our ablation study in Appendix F, including additional results for the effectiveness of the temporal dimension in $E$, comparison of different patch sizes for "EA" and training with different input choices.

\begin{table}[h!]
\centering
\begin{tabular}{c|cccc|cccc} 
\toprule
Model          & $F_E$ & $F_A$ & $F_D$ & $G_P$ & SSIM$\uparrow$            & PSNR$\uparrow$              & FSIM$\uparrow$            & LPIPS$\downarrow$            \\ 
\hline
"No $G_P$"  & +        & +        & +        & -        & 0.5338         & 20.2712          & 0.7399         & 0.4482          \\
"Linear $F$" & *        & -        & -        & +        & 0.4585         & 18.8164           & 0.7006          & 0.4845          \\
"E Only"       & *        & -        & -        & +        & 0.4761          & 19.0881          & 0.7146         & 0.4604          \\
"ED Only"      & +        & -        & +        & +        & 0.5414         & 20.2488          & 0.7392         & 0.4340          \\
"A Only"       & -        & +        & -        & +        & 0.5280           & 20.0312           & 0.7196         & 0.4418           \\
\hline
"EA64"         & +        & +        & -        & +        & \textbf{0.5954} & \textbf{21.2050} & 0.7586         & 0.4053           \\
"EAD"          & +        & +        & +        & +        & 0.5848         & 21.1291          & \textbf{0.7592} & \textbf{0.3985}  \\
\bottomrule
\end{tabular}
\caption{Ablation study on the effects of different components of our model on Texas housing dataset. "+" represents adding certain components, "-" represents removing the components, and "*" represents different configurations from the original setting. See Section \ref{ablation_study} for more details.}
\label{tab:ablation}
\end{table}

\subsection{Human Evaluation for Downstream Applications}
\label{human_eval}

\begin{table}[h!]
\centering
\begin{tabular}{c|cc|cc|cc} 
\toprule
\multirow{2}{*}{Images} & \multicolumn{2}{c|}{$r^2$ with mean count} & \multicolumn{2}{c|}{$r^2$ with median count} & \multicolumn{2}{c}{\% times selected}  \\ 
\cline{2-7}
                    & Buildings    & Pools   & Buildings    & Pools & Similarity  & Realism\\ 
\hline
HR $t'$          & 0.1475    & 0.1009     & 0.1595       & 0.1997   & - & -\\
DBPN                & 0.8785    & 0.0227     & 0.8823       & -0.0640    & 1.75\%    & 1.25\%    \\
cGAN Fusion         & 0.8793    & -0.0707    & 0.9093       & -0.0367     & 45.00\%  & 49.00\%  \\
\hline
Ours (EAD)          & \textbf{0.9174}    & \textbf{0.6158}   & \textbf{0.9298}   & \textbf{0.5953}    & \textbf{53.25\%}    & \textbf{49.75\%} \\
\bottomrule
\end{tabular}
\caption{Human evaluation results on Texas housing dataset.
}
\label{tab:countsrsquared}
\end{table}

Because our goal is to generate realistic and meaningful HR images that can benefit downstream tasks, we also conduct human evaluations to examine the potential of using our models for downstream applications. 
We deploy three human evaluation experiments on Amazon Mechanical Turk to measure the object reconstruction performance, similarity to ground truth HR images, and perceived realism of images generated by different models.

\paragraph{Building and Swimming Pool Count}
Object counting in HR satellite imagery has numerous applications, including environmental regulation~\cite{lee2021scalable},  aid targeting~\cite{sprohnle2017object}, and 
local-level poverty mapping \cite{ayush2020generating}. Therefore, we choose object counting as the primary downstream task for human evaluation. We randomly sample 200 locations in the test set of our Texas housing dataset, and assess the image quality generated under the setting $t'=2018 > t=2016$. Each image is evaluated by 3 workers, and each  worker is asked to count the number of buildings as well as the number of swimming pools in the image. In each location, we select images generated from our model (EAD), cGAN Fusion, and DBPN, as well as the corresponding ground truth HR image $I_{hr}^{(t)}$ from 2016 and HR image $I_{hr}^{(t')}$ from 2018 (denoted as HR $t'$).
We choose buildings and swimming pools as our target objects since both can serve as indicators of regional wealth \cite{ayush2020generating, Carlucci2020} and both occur with high frequency in the areas of interest. Swimming pools are particularly challenging to reconstruct due to their small size and high shape variation, making them an ideal candidate for measuring small-scale object reconstruction performance.

We measure the performance of each setting using the square of Pearson's correlation coefficient ($r^2$) between true and estimated counts, as in previous research~\cite{ayush2020generating}. 
As human-level object detection is still an open problem especially for satellite imagery \cite{Liu2020, Tayara2018}, human evaluation on this task serves as 
an upper bound on the performance of automatic methods  
on this task. 

As shown in Table \ref{tab:countsrsquared}, our model outperforms baselines on both tasks, with the most significant performance advantage in the swimming pool counting task. Note that in rapidly changing environments like our Texas housing dataset, using the HR image of a nearby timestamp $t'$ cannot provide an accurate prediction of time $t$, which indicates the importance of obtaining higher temporal resolution in HR satellite imagery. Our model maintains the best object reconstruction performance among all models experimented, especially for small scale objects.

\paragraph{Similarity to Ground Truth and Image Realism}
Aside from object counting, we also conduct human evaluation on the image sample quality. With 400 randomly selected testing locations in our Texas housing test set, each worker is asked to either select the generated image that best matches a given ground truth HR image, or select the most realistic image from 3 generated images shown in random order. All images are generated under the setting $t' = 2018 > t = 2016$, and we choose the same models as the ones in the object counting experiment. Human evaluation results on image sample quality align with our quantitative metric results. Our model produces the most realistic and accurate images among the compared models. Note that although cGAN Fusion generates realistic images, it fails to maintain structural information accuracy, resulting in lower performance in the similarity to ground truth task.

\subsection{Temporal Extrapolation}
\begin{figure}[h!]
    \centering
    \includegraphics[width = \textwidth]{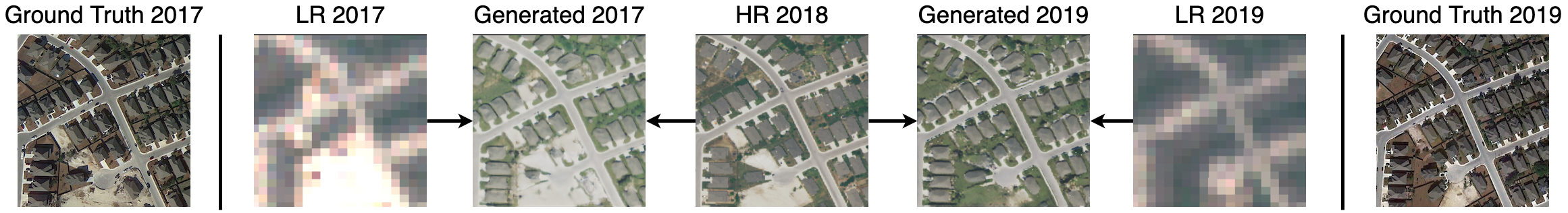}
    \caption{Temporal extrapolation application of our model. HR 2018 is the input NAIP image to both generated images shown in the figure. LR 2017 and LR 2019 are Sentinel-2 images of the same region in 2017 and 2019. Since NAIP imagery is not available in Texas for 2017 and 2019, the ground truth is obtained via Google Earth Pro. Note that the capture dates of the ground truth and the LR images are not perfectly aligned due to a lack of image availability.}
    \label{fig:time}
\end{figure}

Given a HR image at any time and LR images at the desired timestamps, we provide some evidence that our model is able to generate HR images at timestamps unavailable in the training dataset. Figure \ref{fig:time} demonstrates an example of such an application of our model. With the LR images from Sentinel-2, we generate HR images in 2017 and 2019, two years that do not have NAIP coverage in Texas. We compare the generated images with ground truth acquired from Google Earth Pro since the corresponding NAIP images are not available. Although the timestamps of the ground truth and LR images are not perfectly aligned, our generated images still show potential in reconstructing structural information reflected in the ground truth. More rigorous assessment of temporal extrapolation performance requires a more extensive dataset, which we leave to future work.

\section{Conclusion and Statement of Broader Impact}
\label{impact}

We propose a conditional pixel synthesis model that uses the fine-grained spatial information in HR images and the abundant temporal availability of LR images to create the desired synthetic HR images of the target location and time. We show that our model achieves photorealistic sample quality and outperforms competing baselines on a crucial downstream task, object counting.

We hope that the ability to extend access to HR satellite imagery in areas with temporally sparse HR imagery availability will help narrow the data gap between regions of varying economic development and aid in decision making. Our method can also reduce the costs of acquiring HR imagery, making it cheaper to conduct social and economic studies over larger geographies and longer time scales.

That being said,
our method does rely on trustworthy satellite imagery provided by reliable organizations. Just like most SR models, our model is vulnerable to misinformation (e.g. the failure cases presented in Appendix G due to unreliable LR input).
Therefore, we caution against using generated images to inform individual policy (e.g. retroactively applying swimming pool permit fees) or military decisions.
Exploration of performance robustness to adversarial examples is left to future study.
Furthermore, we acknowledge that increasing the temporal availability of HR satellite imagery has potential applications in surveillance.
Finally, we note that object counting performance is measured through human evaluation due to dataset limitations, and we leave measurement of automated object counting performance to future work.
\section{Acknowledgement}
This research was supported in part by NSF (\#1651565, \#1522054, \#1733686), ONR (N00014-19-1-2145), AFOSR (FA9550-19-1-0024), ARO (W911NF-21-1-0125), Sloan Fellowship, HAI, IARPA, and Stanford DDI. This research is based upon work supported in part by the Office of the Director of National Intelligence (ODNI), Intelligence Advanced Research Projects Activity (IARPA), via 2021-2011000004. The views and conclusions contained herein are those of the authors and should not be interpreted as necessarily representing the official policies, either expressed or implied, of ODNI, IARPA, or the U.S. Government. The U.S. Government is authorized to reproduce and distribute reprints for governmental purposes not-withstanding any copyright annotation therein.

\bibliographystyle{plain}
\bibliography{reference}
\newpage
\appendix

\section{Model Details}

We implement $F_E$ and $F_D$ with convolutional layers with kernel size $>1$ with detailed specification below. All convolutional layers are followed by LeakyReLU activation.

$F_A$ is a self-attention module that takes $I_{ne}^{(t)}$ as the input and learns functions $Q, K: \mathbf{R}^{C_{fea} \times H' \times W'} \to \mathbf{R}^{C_{fea}/8 \times H'W'}, V: \mathbf{R}^{C_{fea} \times H' \times W'} \to \mathbf{R}^{C_{fea} \times H'W'}$ and a scalar parameter $\gamma$ to map $I_{ne}^{(t)}$ to $I_{gl}^{(t)}$. Inspired by \cite{sagan, transformer}, $I_{gl}^{(t)} = \gamma * v \textrm{softmax}(k^Tq) + I_{ne}^{(t)}$, where $q = Q(I_{ne}^{(t)}), k = K(I_{ne}^{(t)}), v = V(I_{ne}^{(t)})$, Each entry $\beta_{(x',y'), (x,y)}$ in $k^Tq$ indicates the extent of attention required from $(x,y)$ when extracting features for $(x',y')$. Note that $softmax(k^Tq)v$ needs to be reshaped into $\mathbf{R}^{C_{fea} \times H' \times W'}$ before the addition.

We provide two versions of the image feature mapper. In version "EAD", we use two $3 \times 3$ convolutional layers with stride $=2$ in $F_E$, a single self-attention module in $F_A$ and two $3 \times 3$ transposed convolutional layers with stride $=2$ in $F_D$. In version "EA", we use one linear layer to map the channels from dimension $C$ to $C_{fea}$ and three $3 \times 3$ convolutional layers with stride $=1$ in $F_E$, a single self-attention module in $F_A$ and an identity function in $F_D$. A skip connection is also added between the linear layer in $F_E$ and the output of $F_A$ in "EA".

Style injection is used in all fully-connected layers in $G_p$ and LeakyReLU is also applied after each fully-connected layer except for the layers that map to the output dimension.

\begin{algorithm}[]
\SetAlgoLined
\KwIn{G: Generator, S: patch size, H, W: size of the original image, $\lambda_s$: patch weight}{}
\KwResult{$\hat{I_{hr}^{(t)}} \in \mathbf{R}^{C \times H \times W}$}
    $S_q = S/4$;

    \For{i in \{0, ..., H/S-1\}}{
        \For{j in \{0, ..., W/S-1\}}{
            $\hat{I_{hr}^{(t)}}(X) := G(X, z | I_{lr}^{(t)}, I_{hr}^{(t')})$ where $X = \{(x,y) | iS \leq x \leq (i+1)S, jS \leq y \leq (j+1)S\}$
        }
    }
    \For{i in \{0, ..., H/S-1\}}{
        \For{j in \{0, ..., W/S-2\}}{
            $\hat{I_{temp}^{(t)}}(X') = G(X', z | I_{lr}^{(t)}, I_{hr}^{(t')})$ where $X' = \{(x,y) | iS \leq x \leq (i+1)S, (j+1)S-2S_q \leq y \leq (j+1)S+2S_q\}$
            $\hat{I_{hr}^{(t)}}(X'') += \lambda_s \hat{I_{temp}^{(t)}}(X'')$ where $X'' = \{(x,y) | iS \leq x \leq (i+1)S, (j+1)S-S_q \leq y \leq (j+1)S+S_q\}$
            $\hat{I_{hr}^{(t)}}(X'') /= (1+\lambda_s)$
        }
    }

    \For{i in \{0, ..., H/S-2\}}{
        \For{j in \{0, ..., W/S-1\}}{
            $\hat{I_{temp}^{(t)}}(X') = G(X', z | I_{lr}^{(t)}, I_{hr}^{(t')})$ where $X' = \{(x,y) | (i+1)S-2S_q \leq x \leq (i+1)S+2S_q, jS \leq y \leq (j+1)S\}$
            $\hat{I_{hr}^{(t)}}(X'') += \lambda_s \hat{I_{temp}^{(t)}}(X'')$ where $X'' = \{(x,y) | (i+1)S-S_q \leq x \leq (i+1)S+S_q, jS \leq y \leq (j+1)S\}$
            $\hat{I_{hr}^{(t)}}(X'') /= (1+\lambda_s)$
        }
    }
 \caption{Sliding Window Generation}
 \label{alg:sliding_window}
\end{algorithm}

\subsection{Training and Inference by Patch}
Since HR image data can be high dimensional, it may not be feasible to process the entire $H\times W$ at once especially for the computationally intensive modules like matrix multiplication in the self-attention layer. Hence we utilize the benefit of the spatially bounded image coordinate grid and the conditional independent generation in $G_P$ to perform patch-based training and inference.

Without processing the original $H\times W$ image, the model instead takes a $H''\times W''$ patch, whose coordinate grid can be represented as $P = \{(x,y,t) | h \leq x \leq h + H'', w \leq y \leq w + W''\}$ where $(h,w,t)$ is the top-left corner of the patch, as its input. $H'' | H$ and $W'' | W$. $F$ treats the patch as a complete image and calculates $I_{fea}^{(t)}(x,y)$ for $(x,y) \in P$. While $e_{f_o}$ still learns the Fourier feature for $(x,y,t)$, we grid sample the corresponding parameters in $e_{co}$ so that $e_{co}$ still maintains the original $H\times W$ dimension. Since the calculation in $G_P$ is conditionally independent among pixels given $I_{fea}^{(t)}$ and $E(x,y,t)$, the calculation of $G_P(E(x,y,t), F(I_{cat}^{(t)}), z)$ remains unchanged.

At training time, we randomly crop a $H''\times W''$ patch as the input; at inference time, we use the same $z$ to generate all patches in the same image to maintain style consistency, and a sliding window technique described in Algorithm \ref{alg:sliding_window} to mitigate borderline artifacts.

\section{Implementation Details}
We choose $H = W = 256$, $C = 3$ (the concatenated RGB bands of the input images), $C_{fea} = 256$, $m = 3$, $n = 14$ and  and $\lambda = 100$ for all of our experiments. We use non-saturating conditional GAN loss for $G$ and $R_1$ penalty for $D$, which has the same network structure as the discriminator in \cite{styleganv2, cips}. We train all models using Adam optimizer with learning rate $2\times 10^{-3}, \beta_0 = 0, \beta_1 = 0.99, \epsilon = 10^{-8}$ on NVIDIA Titan XP GPUs. We train each model to convergence which takes around 4-5 days on 1 NVIDIA Titan XP GPU. EAD models can sample 10 images per second and EA64 models can sample 1500 images in around 19 minutes (1.3 image/s). The difference in inference times results from the inference by patch technique described in Appendix A.1.

In our Texas housing experiment, we use time unit $u = 2$ and denote the images from 2016 as $t = 0$ and the images from 2018 as $t = 2$. In our fMoW-Sentinel2 crop field experiment, we use $u = 365$ and denote the starting date of Sentinel-2 imagery, 2015-06-23, as $t = 0$. The temporal coordinates are then normalized with chosen time unit. We implement our code based on the official PyTorch implementation of CIPS \cite{cips}.

We use the same backbone network structures for cGAN Fusion and Pix2Pix with 6-channel input $I_{cat}^{(t)}$. Because neither SRGAN nor DBPN provides $10 \times$ SR, we resize the LR images to $32 \times 32$ and perform $8 \times$ enlargement.

Pixel values are re-scaled to $[0,1]$ during evaluation. We use \cite{iqa}'s implementation for SSIM, FSIM and LPIPS, and implement PSNR based on \cite{pytorch-tools}'s implementation.

\clearpage
\section{Datasets}
\subsection{Texas Housing Dataset}

We collect geo-coordinates for residential houses effectively built between 2014 and 2017 in Texas from the CoreLogic tax and deed database. To make data extraction more efficient, we use DBSCAN \cite{dbscan} to spatially cluster the geo-locations of the houses, and choose four clusters around major metropolitan areas in Texas (Austin/San Antonio, Dallas, Houston, and Waco) which encapsulate most of the datapoints obtained from the CoreLogic database. We then find the latitude and longitude coordinates that bound each cluster and extract rectangles
for both NAIP and Sentinel-2 from Google Earth Engine (GEE) \cite{gorelick2017google} so we can run fast local extraction. The Sentinel-2 rectangles are cloud filtered using the Sentinel-2 Cloud Probability image collection\footnote{Guidance for cloud filtering using the Sentinel-2 Cloud Probability dataset is available \href{https://developers.google.com/earth-engine/tutorials/community/sentinel-2-s2cloudless}{\textcolor{blue}{here}}.}, and the NAIP rectangles are cloud filtered by its data collection process. We choose to obtain images from year 2016 and 2018, which is the range of mutual availability of NAIP and Sentinel-2 in Texas. NAIP coverage of Texas ranges from August to September for both years, so we acquire the corresponding Sentinel-2 image from the same time range.

\begin{figure}[h!]
\begin{subfigure}{.35\textwidth}
  \centering
  \includegraphics[width=.95\linewidth]{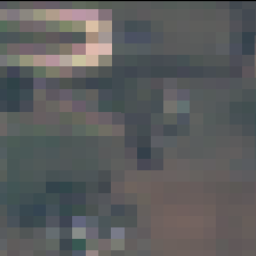}
  \caption{Low resolution image (Sentinel-2)}
\end{subfigure}
\begin{subfigure}{.35\textwidth}
  \centering
  \includegraphics[width=.95\linewidth]{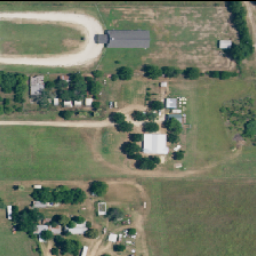}
  \caption{High resolution image (NAIP)}
\end{subfigure}
\caption{\label{fig:naipsentinel}Sentinel-2 and NAIP image pair example.}
\end{figure}

We then extract 4 images for each house (NAIP in 2016, NAIP in 2018, Sentinel-2 in 2016, and Sentinel-2 in 2018). NAIP images have a resolution of 1 meter per pixel, Sentinel-2 images are 10 meters per pixel, and we acquire the RGB bands from both devices. Our dataset consists of 286717 houses and their surrounding neighborhoods. We reserve 14101 houses from 20 randomly selected zip codes as the testing set and use the remaining 272616 houses from the other 759 zip codes as the training set. In total, there are 1146868 multi-resolution images collected from different sensors for our experiment. Table \ref{tab:housedistribution} shows a distribution of houses by year and also by region. Figure \ref{fig:naipsentinel} shows an example of a house captured using this process with NAIP and Sentinel-2.

For both NAIP and Sentinel-2 images, we export from GEE's image pyramid at scale 1, which is equivalent to applying geo-referenced nearest neighbor resampling to the Sentinel-2 images in order to obtain images with the same dimensions as NAIP. Each neighborhood has radius of $0.001$ degrees. The resulting images from both devices have dimensions of around $256 \times 256$.

\begin{table}[h!]
\centering
\begin{tabular}{|c | c | c | c | c | c|}
\hline
Years & Austin & Dallas & Houston & Waco & \textbf{Total} \\
\hline
2014          & 19630 & 24625 & 20327 & 2035 & \textbf{66617}\\
\hline
2015          & 21673 & 27797 & 20347 & 2100 & \textbf{71917}\\
\hline
2016          & 20327 & 30329 & 17741 & 2299 & \textbf{70696}\\
\hline
2017          & 23997 & 32777 & 18312 & 2401 & \textbf{77487}\\
\hline
\textbf{Total}  & \textbf{85627}   & \textbf{115528}  & \textbf{76727}   & \textbf{8835}   & \textbf{286717}\\
\hline
\end{tabular}
\caption{\label{tab:housedistribution}Distribution of collected houses by year and region.}
\end{table}

Sentinel-2 data is provided courtesy of "Copernicus Sentinel data 2015-2020" as outlined by the European Space Agency (ESA), and NAIP data is provided courtesy of U.S. Department of Agriculture (USDA) Farm Production and Conservation - Business Center, Geospatial Enterprise Operations. We have removed all personally identifiable information from our curated dataset.

\label{app:datasets}

\subsection{fMoW-Sentinel2 Crop Field Dataset}
\begin{figure}[h!]
\begin{subfigure}{.35\textwidth}
  \centering
  \includegraphics[width=.95\linewidth]{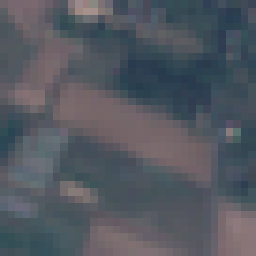}
  \caption{Low resolution image (Sentinel-2)}
\end{subfigure}
\begin{subfigure}{.35\textwidth}
  \centering
  \includegraphics[width=.95\linewidth]{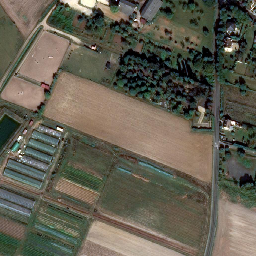}
  \caption{High resolution image (fMoW)}
\end{subfigure}
\caption{\label{fig:fmowsentinel}Sentinel-2 and fMoW image pair example.}
\end{figure}

We derive this dataset from the crop field category of Functional Map of the World (fMoW) dataset \cite{christie2018functional}. We take RGB images from the fMoW crop field object category due to a high likelihood of changes over time compared to other object classes in the fMoW dataset. We pair each fMoW image (0.3m to 1m GSD) with a corresponding lower resolution Sentinel-2 RGB image (10m GSD) captured at the same location and within a 20 day range centered around the fMoW image capture time. We select the least cloudy image from available images within the 20 day capture range using the Sentinel-2 Cloud Probability dataset\footnote{The Sentinel-2 Cloud Probability dataset can be found \href{https://developers.google.com/earth-engine/datasets/catalog/COPERNICUS_S2_CLOUD_PROBABILITY}{\textcolor{blue}{here}}.}. We prune locations with fewer than 2 timestamps, yielding 1752 locations and a total of 4898 fMoW-Sentinel2 pairs. Each location contains between 2-15 timestamps spanning from 2015 to 2017. We reserve 237 locations as the testing set and the remaining 1515 locations as the training set. We resize images to $256 \times 256$ using nearest neighbour interpolation during preprocessing. Figure \ref{fig:fmowsentinel} shows an example of an image pair after data collection and preprocessing.

Sentinel-2 data is provided courtesy of "Copernicus Sentinel data 2015-2020" as outlined by the European Space Agency (ESA), and fMoW images are used under the "Functional Map of the World Challenge Public License"\footnote{The Functional Map of the World Challenge Public License can be found \href{https://github.com/fMoW/dataset/blob/master/LICENSE}{\textcolor{blue}{here}}.}. The dataset does not contain personally identifiable information.

\newpage
\section{Human Evaluation}
We categorize models into three groups: image-to-image translation models, super-resolution models and our models. For human evaluation experiments, we primarily compare  cGAN Fusion, DBPN and our model (EAD) because they achieve the best quantitative metrics in each model category. We conduct all human evaluation under the setting of $t' > t$ on the Texas housing dataset.

\subsection{Building and Swimming Pool Count}
We measure the object reconstruction performance of images at two scales through a task where workers count the number of buildings and swimming pools in a satellite image.
\paragraph{Experiment setup}
200 locations are randomly sampled from the test set, and the corresponding images are collected from the following datasets/models: ground truth HR image $I_{hr}^{(t)}$ from 2016, HR image $I_{hr}^{(t')}$ from 2018 (HR $t'$), cGAN Fusion, DBPN and our model (EAD). 10 images and one vigilance test image are packed into each Human Intelligence Task (HIT) and 3 assignments are requested for each HIT, allowing for comparison of results from different workers.
\paragraph{Instructions full text}
"Please only participate in this HIT if you have normal color vision. The HIT should take approximately 5 minutes to complete. There are 11 trials in the HIT, and each should take 15-30 seconds. You are shown a satellite image that may be real or computer generated. Please count the number of buildings and the number of swimming pools in the image. Please only count buildings and swimming pools that are fully contained within the image, not noticeably clipped off by the edges of the image. Objects that are just barely clipped off (less than 5\%) may be counted. Objects in generated satellite images may look ambiguous. Give your best guess count in these cases. This can be subjective so follow your instincts! You will complete a short practice session (approximately 1 minute) before starting the main task."
\paragraph{Vigilance Tests}
Each HIT contains a vigilance image for which object counts are sufficiently unambiguous. HITs that fail the vigilance test are automatically rejected.
\paragraph{Worker qualification}
Workers are required to have no less than 97\% HIT approval rate and no less than 5000 total approved HITs to qualify for our experiment. Worker uniqueness is used to prevent the same workers from completing more than 3 total HITs for this experiment.
\paragraph{Compensation}
Workers are awarded \$0.80 upon completion of a HIT, which has an expected completion time of 6 minutes, yielding an estimated hourly wage of \$8. We deploy a total of 300 HITs, resulting in a total cost of \$240 for this experiment, not including additional fees.

\begin{figure}[h!]
\begin{subfigure}{.49\textwidth}
  \centering
  \includegraphics[width=.95\linewidth,height=.6\linewidth]{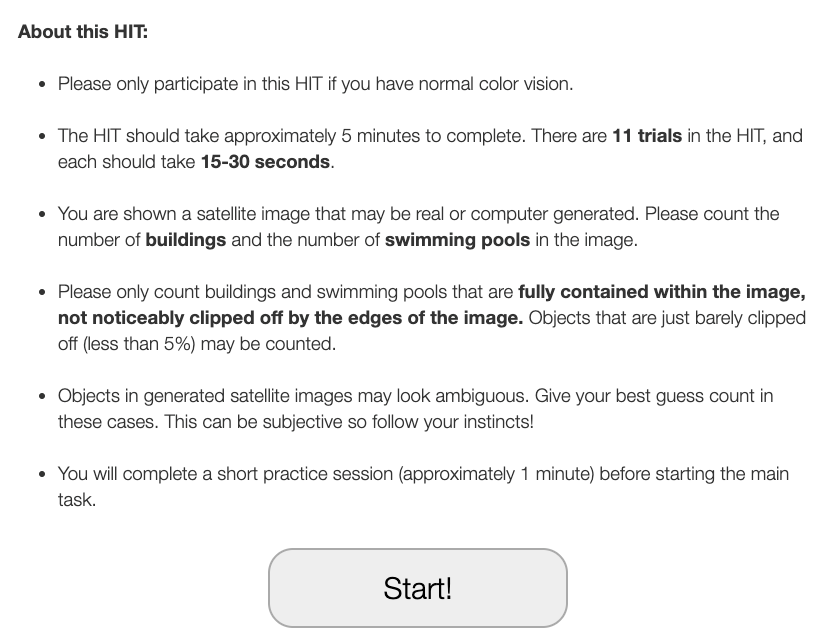}
  \caption{Instructions}
\end{subfigure}
\begin{subfigure}{.49\textwidth}
  \centering
  \includegraphics[width=.95\linewidth,height=.6\linewidth]{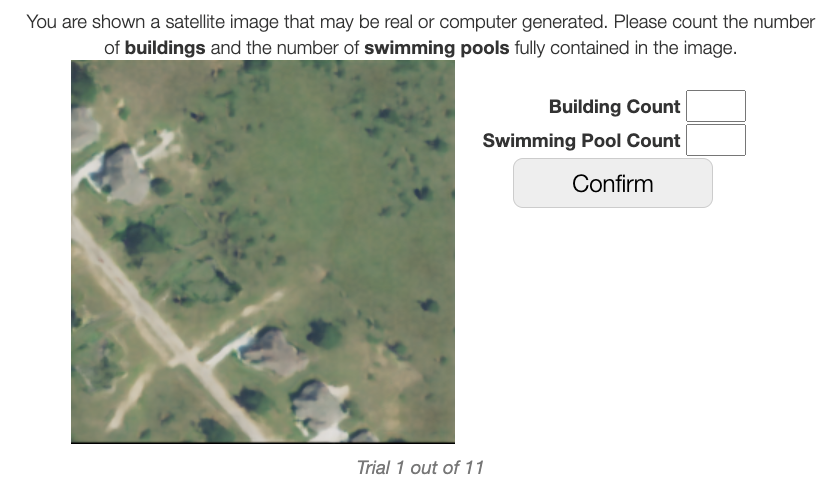}
  \caption{Example Trial}
\end{subfigure}
\caption{Instructions and example trial for the building and swimming pool count task.}
\end{figure}

\subsection{Similarity to Ground Truth}
We measure both structural information accuracy and perceived realism of generated images through a task in which workers select the generated satellite image they observe to best match a real ground truth image.
\paragraph{Experiment setup}
400 locations are randomly sampled from Texas housing dataset test set, and the corresponding images are collected from the following datasets/models: ground truth HR image $I_{hr}^{(t)}$ from 2016, cGAN Fusion, DBPN and our model (EAD). 20 sets of images and two sets of vigilance test images are packed into each HIT and 1 assignment is requested for each HIT.
\paragraph{Instructions full text}
"Please only participate in this HIT if you have normal color vision. The HIT should take 6-7 minutes to complete. There are 22 trials in the HIT, and each should take 12-15 seconds. You are shown a real satellite image and three computer generated satellite images. Please select the generated image that you feel best matches the real image based on image accuracy and realism. Image accuracy refers to the degree to which ground level structural information (e.g. buildings, trees, roads) matches the real image. Image realism refers to the degree that the generated image looks like a real satellite image. You will complete a short practice session (less than 1 minute) before starting the main task."
\paragraph{Vigilance Tests}
Each HIT contains a similarity trial for which the best matching image is sufficiently unambiguous. HITs that fail the vigilance test are automatically rejected.
\paragraph{Worker qualification}
Workers are required to have no less than 97\% HIT approval rate and no less than 5000 total approved HITs to qualify for our experiment. Worker uniqueness is used to prevent the same workers from completing more than 1 HIT for this experiment.
\paragraph{Compensation}
Workers are awarded \$1.00 upon completion of a HIT, which has an expected completion time of 6 minutes, yielding an estimated hourly wage of \$10. We deploy a total of 20 HITs, resulting in a total cost of \$20 for this experiment, not including additional fees.

\begin{figure}[h!]
\begin{subfigure}{.49\textwidth}
  \centering
  \includegraphics[width=.95\linewidth,height=.45\linewidth]{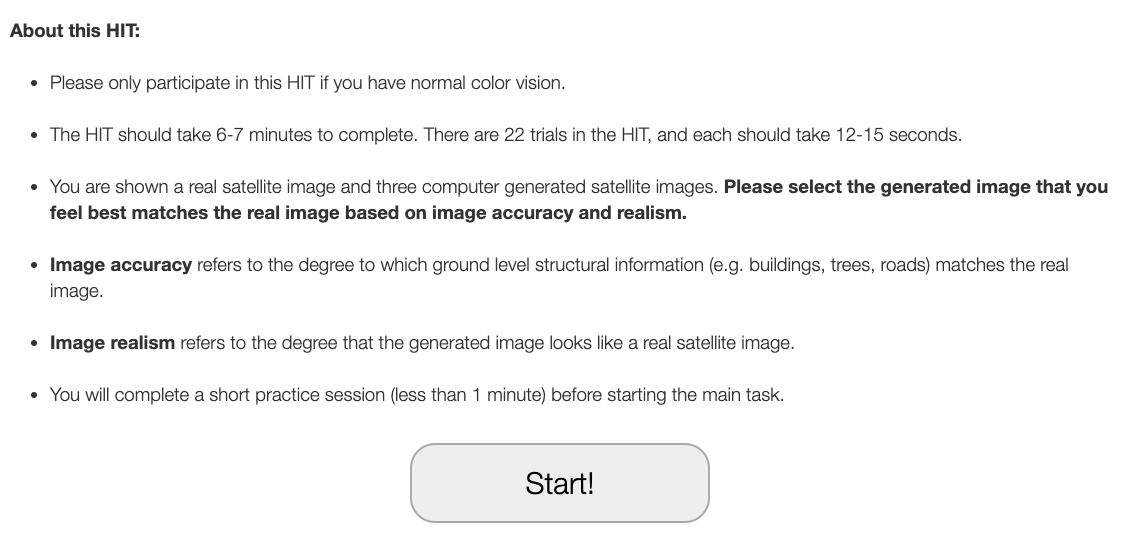}
  \caption{Instructions}
\end{subfigure}
\begin{subfigure}{.49\textwidth}
  \centering
  \includegraphics[width=.95\linewidth,height=.45\linewidth]{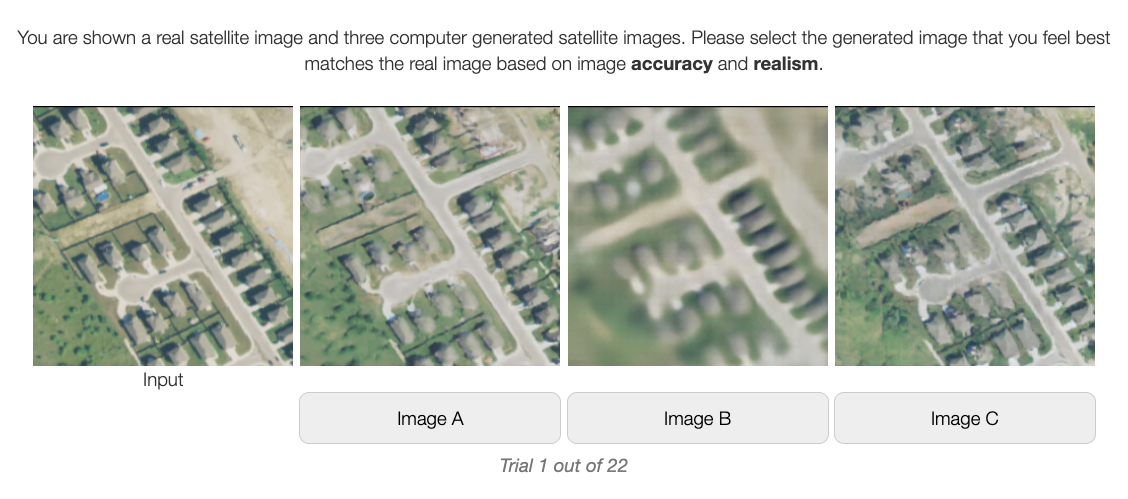}
  \caption{Example Trial}
\end{subfigure}
\caption{Instructions and example trial for the similarity to ground truth task.}
\end{figure}

\subsection{Image Realism}
We measure the perceived realism of generated images through a task in which workers select the generated satellite image they observe to look the most realistic. Workers are provided with real satellite images to examine in the beginning of the task.
\paragraph{Experiment setup}
400 locations are randomly sampled from Texas housing test set, and the corresponding images are collected from the following models: cGAN Fusion, DBPN and our model (EAD). 20 sets of images and two sets of vigilance test images are packed into each HIT and 1 assignment is requested for each HIT.
\paragraph{Instructions full text}
"Please only participate in this HIT if you have normal color vision. The HIT should take 6-7 minutes to complete. There are 22 trials in the HIT, and each should take ~15 seconds. You are shown three computer generated satellite images. Please select the generated image that most resembles a high quality real satellite image. Please pay attention to the details of objects. You will complete a short practice session (approximately 1 minute) before starting the main task."
\paragraph{Vigilance Tests}
Each HIT contains a realism trial for which the most realistic satellite image is sufficiently unambiguous. HITs that fail the vigilance test are automatically rejected.
\paragraph{Worker qualification}
Workers are required to have no less than 97\% HIT approval rate and no less than 5000 total approved HITs to qualify for our experiment. Worker uniqueness is used to prevent the same workers from completing more than 1 HIT for this experiment.
\paragraph{Compensation}
Workers are awarded \$1.00 upon completion of a HIT, which has an expected completion time of 6 minutes, yielding an estimated hourly wage of \$10. We deploy a total of 20 HITs, resulting in a total cost of \$20 for this experiment, not including additional fees.

\begin{figure}[h!]
\begin{subfigure}{.49\textwidth}
  \centering
  \includegraphics[width=.95\linewidth,height=.53\linewidth]{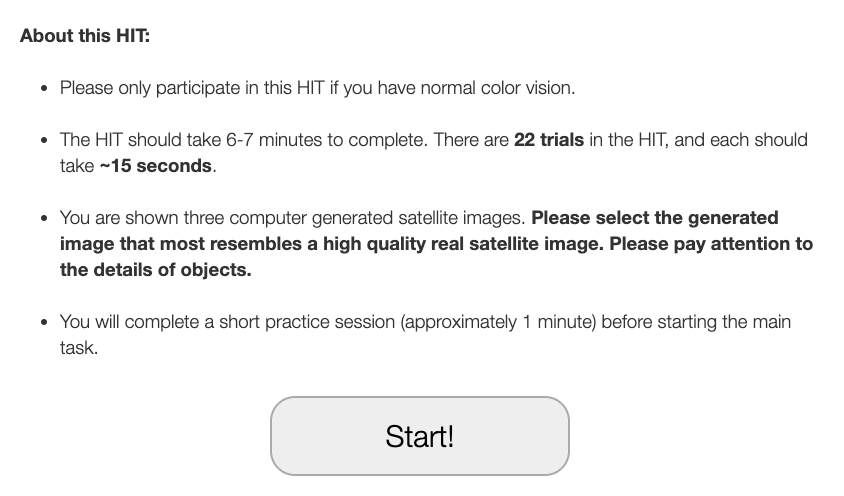}
  \caption{Instructions}
\end{subfigure}
\begin{subfigure}{.49\textwidth}
  \centering
  \includegraphics[width=.95\linewidth,height=.53\linewidth]{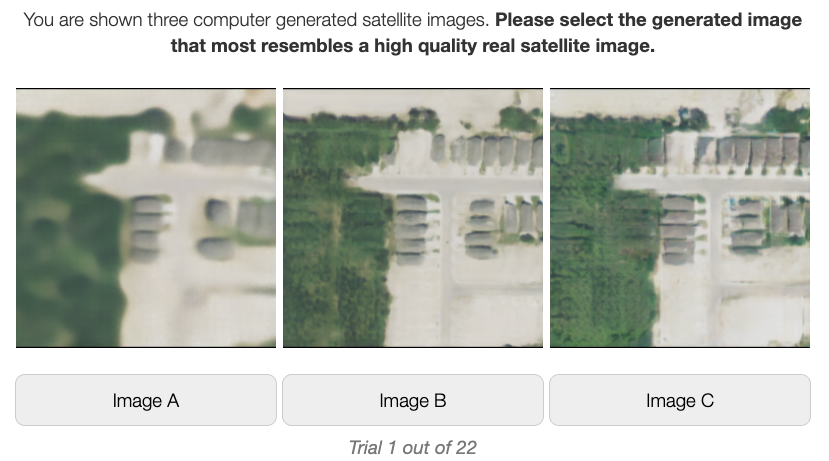}
  \caption{Example Trial}
\end{subfigure}
\caption{Instructions and example trial for the image realism task.}
\end{figure}

\subsection{Additional Statistical Results}
\begin{figure}[h!]
\begin{subfigure}{.45\textwidth}
  \centering
  \includegraphics[width=.85\linewidth]{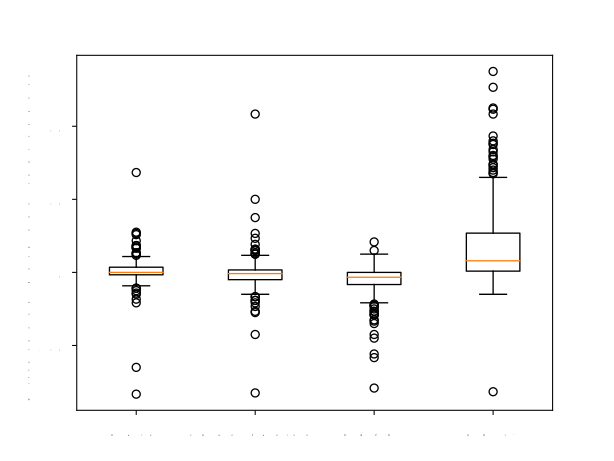}
  \caption{Difference in Building Counts}
\end{subfigure}
\begin{subfigure}{.45\textwidth}
  \centering
  \includegraphics[width=.85\linewidth]{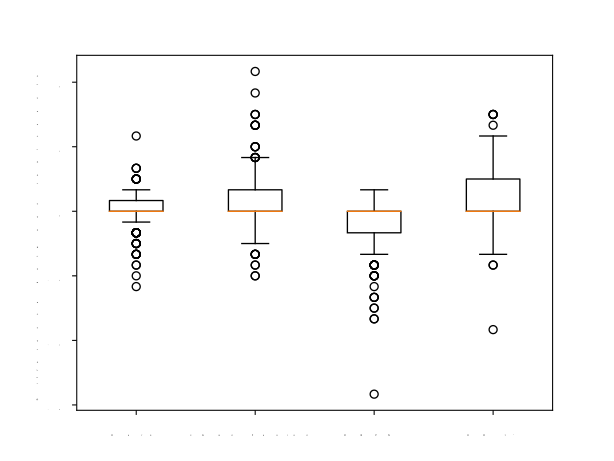}
  \caption{Difference in Pool Counts}
\end{subfigure}
\caption{Box plots of differences in mean object count for each dataset/model compared to the mean object counts yielded by the ground truth images.}
\label{fig:countsbox}
\end{figure}

Figure \ref{fig:countsbox} shows box plots of differences between mean object counts in each location for images generated by each model compared to the ground truth images. We use $IQR = Q_3 - Q_1$ and plot non-outliers in $[Q_1 - 1.5\times IQR, Q_3 + 1.5\times IQR]$. We remind the reader that object counts for ground truth images are also generated by human workers through the same experiment. Compared to our model, cGAN Fusion yields greater variation in object count difference and has a tendency to hallucinate swimming pools. Images generated by DBPN often yield under-counted buildings and swimming pools, likely due to image blurriness and signal limitations from LR input. We note that building counts obtained from the HR image at a future time $t'$ are frequently higher than building counts obtained from the HR image at a past time $t$, sometimes significantly so. This can be attributed to the rapidly changing environments and the presence of new house constructions in our dataset. Count differences derived from images generated by our model exhibit smaller variance and less data skew when measured against counts derived from the ground truth images.

\clearpage
\section{Additional Generation Results}

\begin{figure}[h!]
    \centering
    \includegraphics[width = \textwidth]{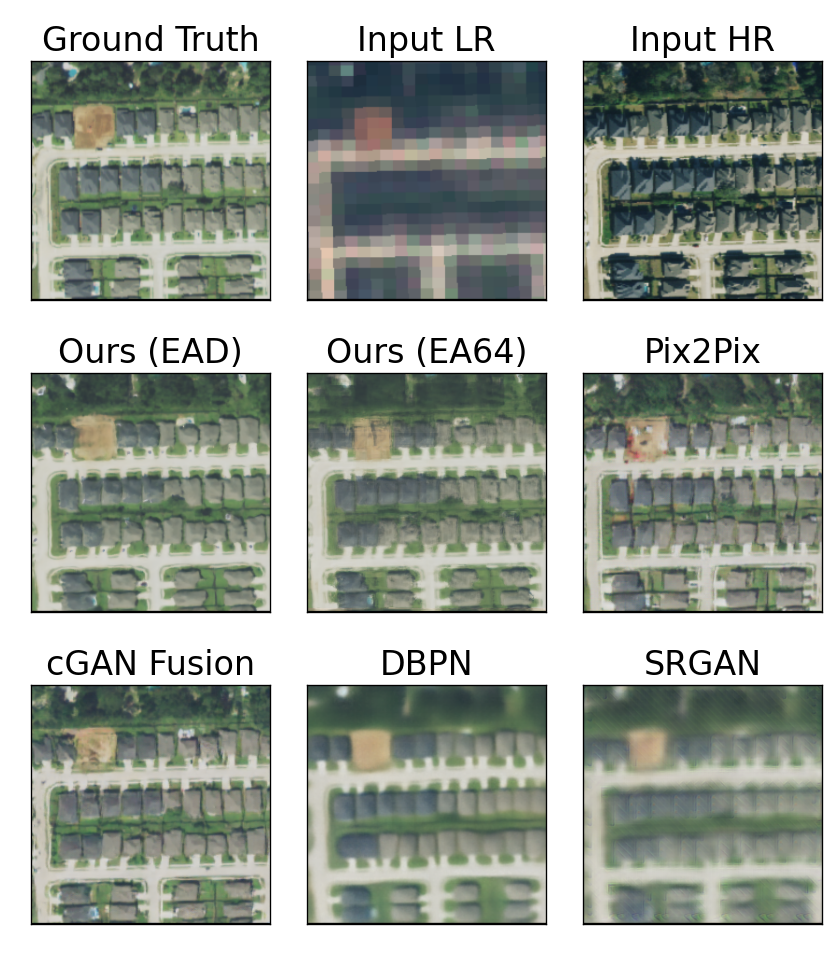}
    \caption{Additional samples from all models on the Texas housing dataset with setting $t' > t$.}
\end{figure}

\begin{figure}[h!]
    \centering
    \includegraphics[width = \textwidth]{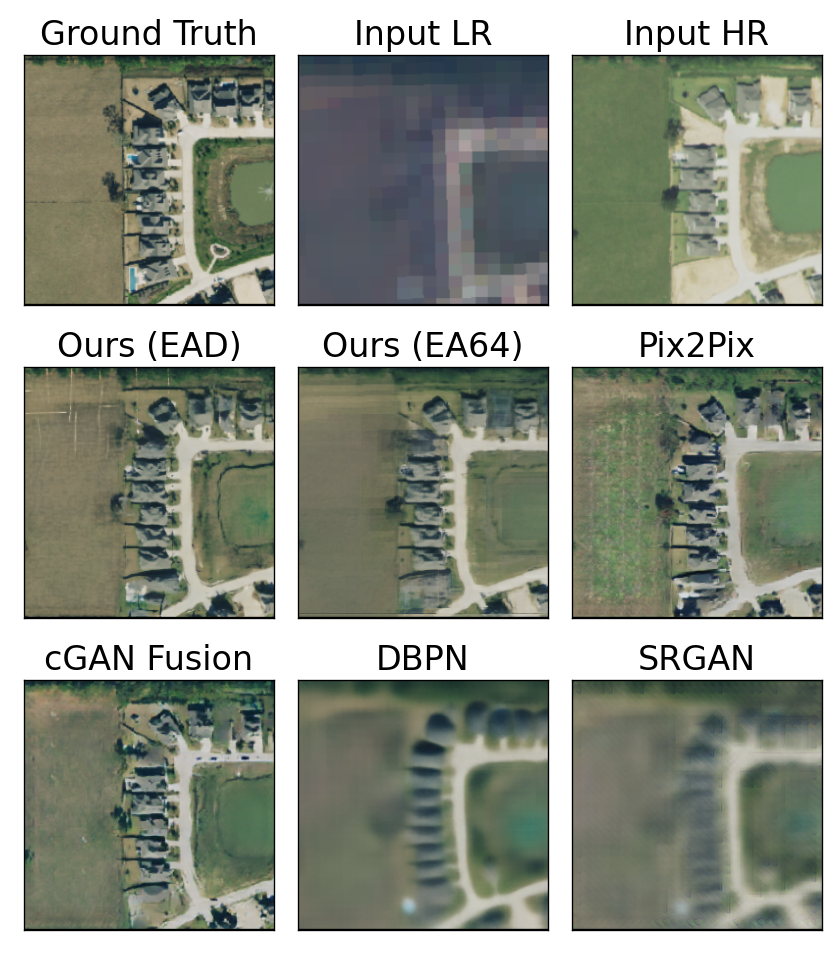}
    \caption{Additional samples from all models on the Texas housing dataset with setting $t' < t$.}
\end{figure}

\begin{figure}
    \centering
    \includegraphics[width = \textwidth]{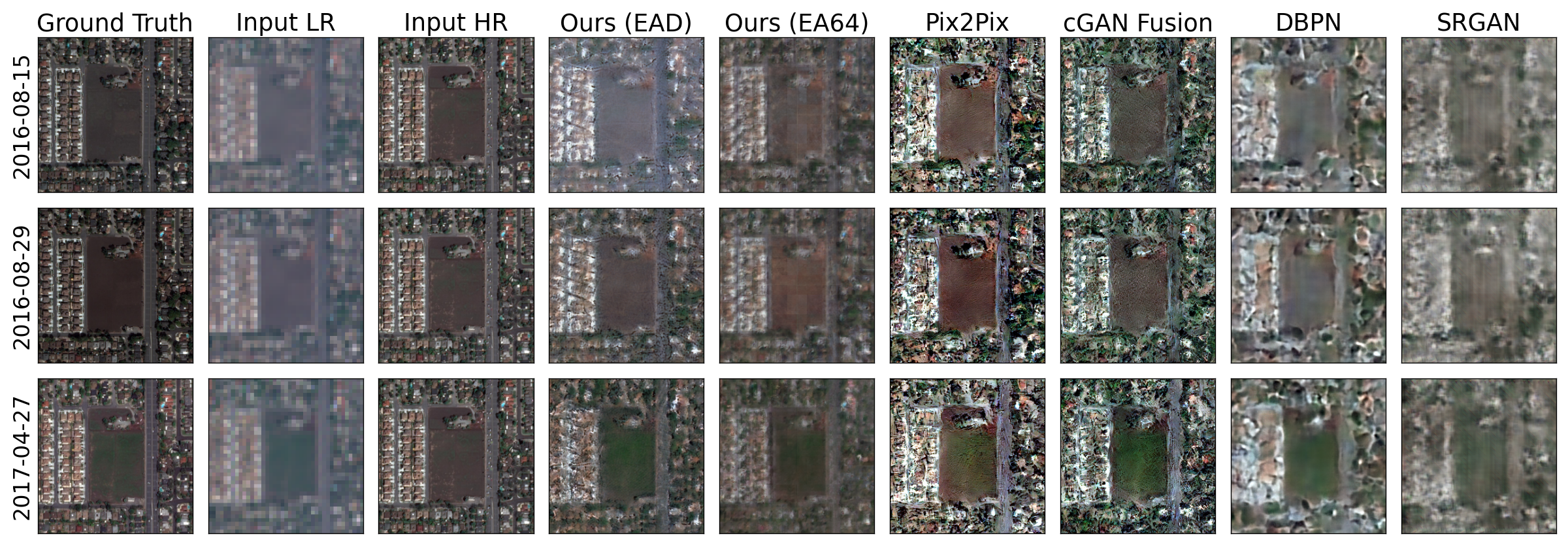}
    \caption{Additional samples from all models on the fMoW-Sentinel2 crop field dataset.}
\end{figure}

\clearpage
\section{Additional Ablation Study}
In this section, we perform an additional ablation study with qualitative examples and quantitative analysis.

\begin{figure}[h!]
    \centering
    \includegraphics[width = 0.95\textwidth]{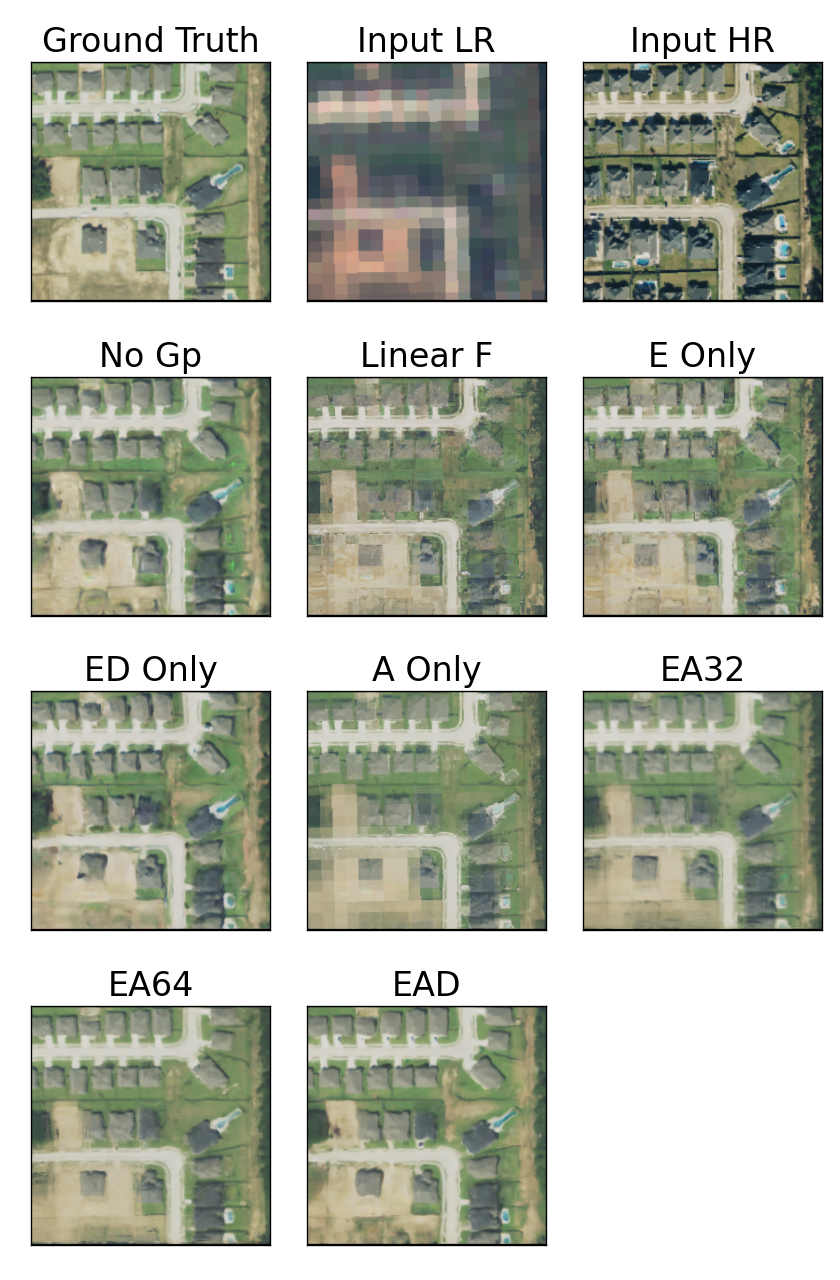}
    \caption{Samples from all different configurations of our model on the Texas housing dataset with setting $t' > t$.}
    \label{fig:ablation}
\end{figure}

\paragraph{Model Configurations} Figure \ref{fig:ablation} shows an example of samples from different configurations of our model mentioned in Section 5.4. From the samples in Figure \ref{fig:ablation}, we observe that although our model still produces convincing results without $G_P$, it suffers from the same high saturation visual artifacts as Pix2Pix and cGAN Fusion. Without deep network structures in $F$, while faithful to the LR image, our model fails to generate realistic results. Without $F_E$ and $F_D$, our model shows obvious checkerboard artifacts. After removing $F_A$, our model generates inconsistent shapes for objects in the image.

\begin{figure}[h!]
\begin{subfigure}{.5\textwidth}
  \centering
  \includegraphics[width=.8\linewidth]{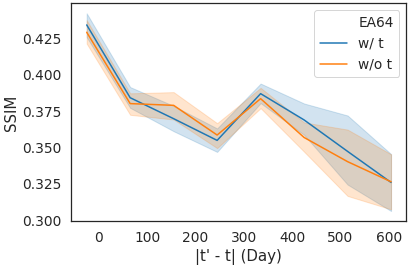}
  \caption{SSIM Results}
  \label{fig:sfig1}
\end{subfigure}%
\begin{subfigure}{.5\textwidth}
  \centering
  \includegraphics[width=.8\linewidth]{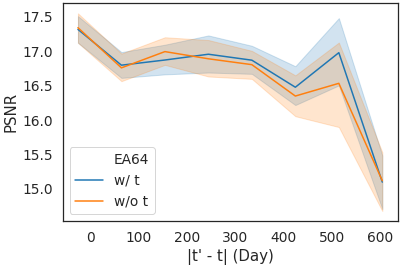}
  \caption{PSNR Results}
  \label{fig:sfig2}
\end{subfigure}
\begin{subfigure}{.5\textwidth}
  \centering
  \includegraphics[width=.8\linewidth]{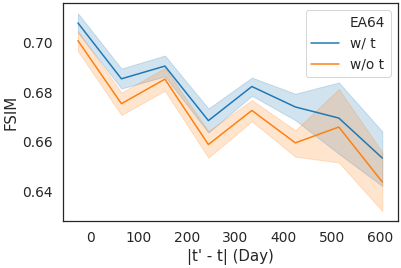}
  \caption{FSIM Results}
  \label{fig:sfig2}
\end{subfigure}%
\begin{subfigure}{.5\textwidth}
  \centering
  \includegraphics[width=.8\linewidth]{images/lpips_t_1.png}
  \caption{LPIPS Results}
  \label{fig:sfig2}
\end{subfigure}
\caption{Ablation study on learning the time dimension in our model (EA64) using fMoW-Sentinel2 crop field dataset.}
\label{fig:t_ablation2}
\end{figure}

\paragraph{Temporal Encoding} We conduct an ablation study on learning the time dimension in $E$ using "EA64" on the fMoW-Sentinel2 crop field dataset. Figure \ref{fig:t_ablation2} demonstrates a comparison of all metrics between models with and without learning the time component in $E$. We group all possible values of $|t'-t|$ in the test set with bin width of $90$ days, and plot the mean of each bin and error bars with $75\%$ confidence level.

As we observe in the figure, the performance of EA64 with time embeddings shows advantages in most metrics as the time difference between capture dates of the target and reference HR images increases, which suggests benefits in real life settings where HR imagery is often captured with long visiting cycles.

Using permutation test with differences between means to test the statistical significance, we verify our observation with the null hypothesis $H_0: f(X_w) \leq f(X_{wo})$ where $f$ is a choice of metrics where higher values indicate better models. Here $f$ can be SSIM, PSNR, FSIM or -LPIPS compared to the ground truth. $X_w$ and $X_{wo}$ represent the generated images with and without learning the time dimension respectively. FSIM and LPIPS achieve p-values 0.002 and 0 respectively for the null hypothesis. PSNR achieves p-value=0.09 when |t’-t|>300 days, which shows that the advantage of the time-embedded model increases when the difference between $t'$ and $t$ increases.

We also report average quantitative results for both models and observe that the model with temporal information outperforms in all metrics on the fMoW-Sentinel2 crop field dataset. Hence we can conclude that learning the time dimension is crucial for our model performance.

\begin{table}[]
\begin{tabular}{c|cccc}
\toprule
Model & SSIM$\uparrow$           & PSNR$\uparrow$             & FSIM$\uparrow$            & LPIPS$\downarrow$           \\ \hline
w/o t & 0.3844          & 16.8075          & 0.6774          & 0.5380          \\
w/ t  & \textbf{0.3905} & \textbf{16.8879} & \textbf{0.6827} & \textbf{0.5197} \\ \bottomrule
\end{tabular}
\caption{Quantitative results of ablation study on learning the temporal information on the fMoW-Sentinel2 crop field dataset. "w/ t" represent EA64 model with time dimension in $E$, and "w/o t" represent the same model without time dimension in $E$. As shown here, adding the temporal dimension substantially benefits model performance.}
\end{table}

\paragraph{Patch Size} We also investigate a setting of "EA" with patch size $32$ denoted as "EA32", with quantitative results presented in Table \ref{tab:ea32}. The model performance is degraded due to the reduction of the global view of $F_A$. We hypothesize that increasing the patch size can benefit the performance. However, further study is required as it is not feasible to train patch sizes that are significantly larger than $64$ on our devices.

\begin{table}[h!]
\centering
\begin{tabular}{c|cccc|cccc}
\toprule
Model          & $F_E$ & $F_A$ & $F_D$ & $G_P$ & SSIM$\uparrow$     & PSNR$\uparrow$    & FSIM$\uparrow$    & LPIPS$\downarrow$\\
\hline
EA32         & +(32)        & +(32)        & -        & +        & 0.5767         & 20.8409           & 0.7480           & 0.4254          \\
\hline
EA64         & +(64)        & +(64)        & -        & +        & \textbf{0.5954} & \textbf{21.2050} & \textbf{0.7586}         & \textbf{0.4053}           \\
\bottomrule
\end{tabular}
\caption{Ablation study on the patch sizes for patch-based generation on the Texas housing dataset. $+(S)$ denotes using size $S$ patches during training and inference.}
\label{tab:ea32}
\end{table}

\paragraph{Input Source} Here we provide an experiment using our model with (1) two additional LR image inputs from different time steps (Sentinel-2 images from 2017 and 2019) and (2) no HR reference input (as in standard SR approaches) in the Texas housing dataset.

We do not observe improvements from the "Multiple LR" setting, which agrees with our claim that given a LR image at the target time, other LR views from different time steps (in the past or future) provide little or no additional information. As shown for the "No HR t'" setting in the table below, our approach also has clear advantages over the standard SR setting (with no HR reference input). It is also worth noting that the LR images in our experiments are from real LR devices, which is different from synthetic LR images created by downsampling used in many SR benchmarks. Leading standard SR methods such as DBPN and SRGAN do not perform well as shown in the experiments in Section 5.

\begin{table}[h!]
\begin{tabular}{c|cccc|cccc}
\toprule
\multirow{2}{*}{Model} & \multicolumn{4}{c|}{t' > t}                                 & \multicolumn{4}{c}{t' < t}                                     \\ \cline{2-9}
                       & SSIM$\uparrow$            & PSNR$\uparrow$             & FSIM$\uparrow$            & LPIPS$\downarrow$           & SSIM$\uparrow$            & PSNR$\uparrow$             & FSIM$\uparrow$            & LPIPS$\downarrow$           \\ \hline
Multiple LR            & 0.6266          & 22.0939          & 0.7814          & 0.3873          & 0.5023          & 19.5255          & 0.7198          & 0.4428          \\
No HR t'               & 0.4731          & 19.9537          & 0.7047          & 0.4991          & 0.3535          & 17.4470          & 0.6453          & 0.5400          \\ \hline
Ours                   & \textbf{0.6470} & \textbf{22.4906} & \textbf{0.7904} & \textbf{0.3695} & \textbf{0.5225} & \textbf{19.7675} & \textbf{0.7280} & \textbf{0.4275} \\ \bottomrule
\end{tabular}
\caption{Ablation study on different input choices on the Texas housing dataset. "Multiple LR" represents the setting that uses more than one LR images from different time steps as inputs, and "No HR t'" represents the setting that uses no HR reference image from a different time and only takes the LR image as the input, as in the standard SR approaches. Our approach outperforms both models with the EAD configuration.}
\label{tab:input_ablation}
\end{table}

Our method can also be easily extended to include additional bands such as NIR by changing the number of input channels. We choose RGB bands in this work because they are commonly available in remote sensing devices and they are sufficient for our target tasks. Thus we see our results as a lower bound of what can be achieved, and enhancements using additional bands are left to future study.

\clearpage
\section{Failure Cases}
\begin{figure}[h!]
    \centering
    \includegraphics[width = \textwidth]{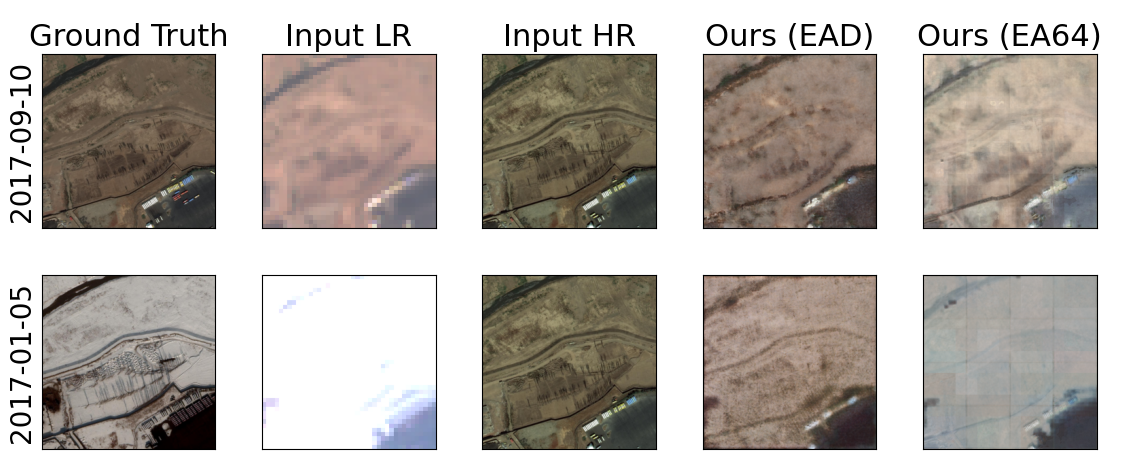}
    \caption{Failure cases when using fMoW-Sentinel2 crop field dataset with input HR captured on 2016-09-29. With this edge case LR input (extreme snow reflection), our EAD model fails to generate accurate color and our EA64 model exhibits obvious checkerboard artifacts.}
    \label{fig:fail_fmow}
\end{figure}

In this section, we analyze failure cases of our model. Figure \ref{fig:fail_fmow} is an example where we observe unsatisfactory generation quality from our model. In some edge cases, such as extreme snow reflection in the LR input image, our model is unable to successfully reconstruct an accurate image of the ground truth. In this case, our EAD setting does not obtain sufficient color accuracy. Our EA64 setting is able to generate more accurate color details for the 2017-01-05 image, but it synthesizes obvious checkerboard artifacts and is therefore less realistic to human perception.

\end{document}